\documentclass{article}
\usepackage{caption}

\usepackage[main, final]{neurips_2026}

\usepackage[utf8]{inputenc} 
\usepackage[T1]{fontenc}    
\usepackage{hyperref}       
\usepackage{url}            
\usepackage{booktabs}       
\usepackage{amsfonts}       
\usepackage{nicefrac}       
\usepackage{microtype}      
\usepackage{graphicx}
\usepackage{amsmath,amssymb,amsthm}
\usepackage{mathrsfs}
\usepackage{booktabs}
\usepackage{makecell}
\usepackage{multirow}
\usepackage[table]{xcolor}
\usepackage{soul}

\usepackage{algorithm}
\usepackage{algpseudocode}

\definecolor{bestred}{RGB}{220,55,55}
\definecolor{secondblue}{RGB}{45,110,220}
\definecolor{thirdgreen}{RGB}{45,150,85}
\definecolor{oursgray}{RGB}{232,232,232}

\newcommand{\best}[1]{\textcolor{bestred}{\textbf{#1}}}
\newcommand{\second}[1]{\textcolor{secondblue}{\textbf{#1}}}
\newcommand{\third}[1]{\textcolor{thirdgreen}{\textbf{#1}}}

\title{Beyond Spatial-Domain Supervision: A Relation Constrained Space for Multi-Modal Image Fusion}

\author{%
  Zeyu Wang \\
  \makebox[0.42\textwidth][c]{College of Computer Science and Engineering}\\
  Dalian Minzu University\\
  \texttt{20231578@dlnu.edu.cn}\\
  \And
  Mingyu Ge \\
  \makebox[0.42\textwidth][c]{College of Computer Science and Engineering}\\
  Dalian Minzu University\\
  \texttt{2019082204@stu.dlnu.edu.cn}\\
  \AND
  Haiyu Song$^{*}$\\
  \makebox[0.42\textwidth][c]{College of Computer Science and Engineering}\\
  Dalian Minzu University\\
  \texttt{shy@dlnu.edu.cn}
  \And
  Haoran Duan\thanks{Corresponding authors.}\\
  \makebox[0.42\textwidth][c]{Department of Automation}\\
  Tsinghua University\\
  \texttt{haoran.duan@ieee.org}
}

\begin{document}

\maketitle
\begin{abstract}
Multi-modal image fusion (MMIF) aims to form a single image by integrating shared information, preserving complementary cues, and coordinating cross-modal conflicts across modalities. However, due to the absence of ground-truth fused images, existing MMIF supervision commonly uses spatial-domain sources or gradient variants as surrogate ground truths, making the supervision mechanism inherently misaligned with the goal of MMIF and causing pixel-level compromise or modality bias. To address this, we propose a relation-constrained supervision paradigm that moves fusion supervision from the spatial domain to a learned relation space. Instead of forcing the fused image to approximate the sources, we use frozen pretrained representation models as information providers and design a learnable feature adapter to align heterogeneous DINO and CLIP features into a unified supervision space. The adapter infers three relation parameters, namely sharedness, dominance, and coordination radius, which define three losses corresponding to the MMIF's goal. To make this space reliable, we devise a self-supervised contrastive ranking objective tailored to the adapter and couple it with the fusion network through alternating optimization. Extensive experiments show that the proposed supervision space yields significant gains regardless of which mainstream backbone the fusion network adopts, offering a supervision paradigm better aligned with the goal of MMIF. Code: github.com/GMY628/RCS-Fusion.
\end{abstract}

\section{Introduction}
Multi-modal Image Fusion (MMIF) combines complementary information from different modalities into one image~\cite{maskdifuser, c2rf, zhang2025omnifuse}. Representative tasks include infrared visible fusion~\cite{liu2024infrared,textdifuse}, which integrates thermal cues with visible textures, and medical image fusion, which combines anatomical and functional information. MMIF also benefits the object detection~\cite{liu2025dcevo} and medical segmentation~\cite{mtgfusion}.

Recent MMIF methods have advanced architectures, cross-modal interaction, learning strategies, and loss design~\cite{wang2025efficient,li2026uncertainty,liu2024promptfusion,CLDyN}. Despite this, due to the absence of ground-truth (GT) fused images, existing methods typically treat spatial-domain source images or their gradient variants as surrogate targets for loss computation. \textbf{This practice has become standard, but introduces a fundamental problem: the supervision mechanism is inherently misaligned with the goal of MMIF}. 
\begin{figure*}[t]
 \centering
  \includegraphics[width=1\textwidth]{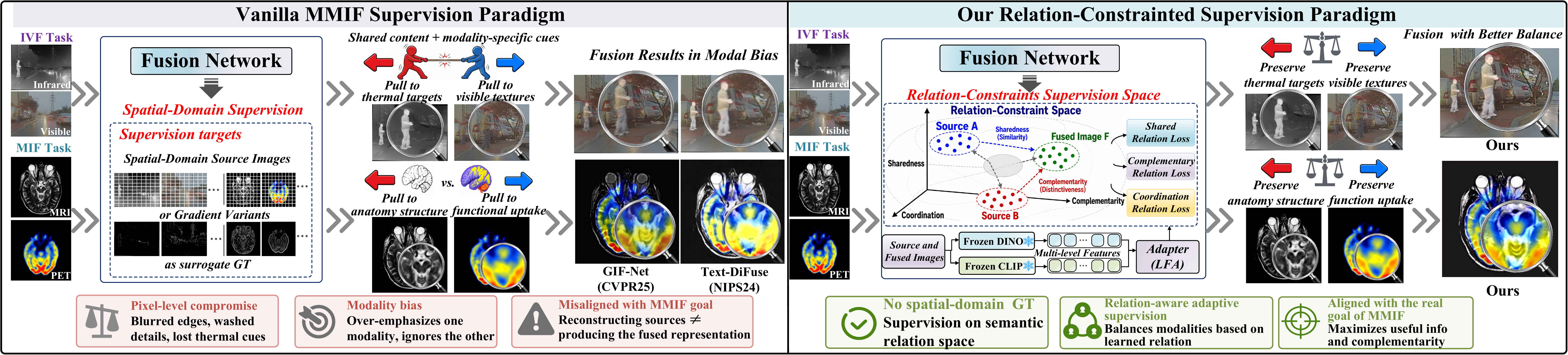}
  \caption{Vanilla MMIF Supervision Paradigm \emph{vs} Our Paradigm. Our method constructs a relation-constraint supervision space tailored for MMIF, leading to better modality balance and fusion quality.}
  \label{first_figure}
\end{figure*}

The goal of MMIF is to integrate shared information, preserve modality specific complementary cues, and coordinate conflicts between heterogeneous observations, rather than reconstruct any individual source image. Source images should therefore act as information providers instead of reconstruction targets. Once they are treated as simultaneous GT, the model is forced to satisfy competing constraints from different modalities. This leads to either a pixel-level compromise or a biased result dominated by one modality, as shown in Fig.~\ref{first_figure}. More importantly, complementary cues, such as infrared thermal targets and visible fine textures, are converted into conflicting supervisory signals.

A natural solution is to move supervision beyond raw spatial domain source images. Recent pretrained representation models~\cite{oquab2023dinov2, CLIP} provide rich visual features that encode structural and semantic cues. Nevertheless, directly using these features does not solve the problem. Features from different pretrained models are heterogeneous and cannot naturally form a unified supervision space. More critically, ideal MMIF supervision requires not only strong representation capacity, but also explicit relation modeling between modalities: what information is shared, what information is complementary, and how conflicting cues should be coordinated. Without such relation modeling, pretrained features may enrich the representation, but supervision remains insufficiently aligned with fusion.

Motivated by this observation, we propose a relation-constrained supervision paradigm for MMIF. Instead of using source images as crude surrogate ground truths, we construct an adaptive supervision space from frozen pretrained representation models and constrain it according to cross-modal relations. Our core idea is that a proper fusion supervision space should be defined by three fundamental relations: sharedness, complementarity, and coordination (details analysis see Appendix~\ref{app:three_relations}). 
Sharedness retains information consistently supported by multiple sources. Complementarity preserves modality specific cues that are informative but weak or absent in the other modality. Coordination prevents conflicting responses from driving the fused result excessively toward either source. Together, these relations provide a supervision principle that matches the intrinsic objective of MMIF.

To implement this principle, we design a learnable feature adapter (LFA) that bridges heterogeneous pretrained features and the proposed supervision space. LFA consists of a Feature Alignment Module and a Relation Reasoning Module. The former projects intermediate features from frozen DINO and CLIP into a unified feature space. Within this space, we tailor three losses, namely sharedness, complementarity, and coordination losses, to supervise fusion. The latter predicts three relation parameters that adaptively weight these losses, allowing supervision to reflect the cross-source relationship of each image pair rather than fixed manual coefficients.

Besides, training the adapter is critical yet challenging. An unreliable adapter would produce misleading supervision and thus undermine any fusion model trained with it. However, no ready made training set or established strategy exists for learning such a relation-aware adapter, creating a circular dependency between adapter learning and fusion supervision. We therefore tailor a self-supervised contrastive training scheme to the adapter and couple it with alternating optimization. Specifically, we pretrain the adapter with a ranking loss on relation scores computed from its outputs, so that matched pairs (positive pairs) score higher than unmatched ones (negative pairs). This enables LFA to learn cross-modal relation estimation without explicit relation annotations. Before alternating optimization, the fusion network is first trained with a conventional fusion loss to reach a baseline level of fusion performance. We then alternate between freezing the adapter to supervise the fusion network with relation losses and freezing the fusion network to train the adapter with fused images under the same ranking objective, encouraging each fused image to score higher with its matched sources than with mismatched ones. This strategy enables effective training of both the adapter and the fusion network. Our contributions are summarized as follows:\\ 
$\bullet$We revisit MMIF supervision from the objective of fusion and propose a relation-constrained paradigm beyond spatial domain source images as surrogate ground truths.\\$\bullet$We define the supervision space through sharedness, complementarity, and coordination, aligning supervision with the requirements of fusion.\\ 
$\bullet$We design a learnable feature adapter that aligns heterogeneous pretrained features and predicts relation parameters to automatically construct the supervision space.\\ $\bullet$We propose a self-supervised relation ranking strategy and an alternating optimization scheme to jointly train the adapter and fusion network, forming a bidirectional optimization mechanism.

\vspace{-2mm}
\section{Related Work}
\textbf{Supervision Paradigms for MMIF.} Recent MMIF research has progressed from reconstruction-based frameworks to recent Transformer, diffusion, and language-guided models~\cite{zhao2020didfuse,li2026drfusion,zhao2024image}. Because no physical ground-truth fused image exists, their performance remains highly dependent on supervision loss design~\cite{bai2025refusion,bai2025task}. Existing methods mainly follow source-preservation, perceptual or adversarial, task-driven, and semantic or language-guided paradigms~\cite{liu2025dcevo, sage, textdifuse}. However, they still use spatial-domain source images as GT or proxy targets for loss computation. This misaligns optimization with the objective of MMIF and often causes pixel-level compromise or modality bias. In contrast, our method abandons this long-standing practice and uses powerful frozen pretrained representation models to build a supervision paradigm better aligned with the goal of MMIF.

\textbf{Pretrained Visual Knowledge for MMIF.} Recent advances in pretrained representation models have made large-scale visual knowledge a promising resource for image fusion. Models such as CLIP and DINOv2 learn rich structural and semantic cues from massive data, and their representations have shown strong transferability in low-level vision tasks \cite{xu2024boosting,cheng2024transfer,li2024contourlet,li2025difiisr}. In image fusion, recent studies have explored knowledge-aware or representation-driven fusion through semantic guidance \cite{zhang2024mrfs,bai2025task,li2026unifusion}, text descriptions \cite{yi2024text,cao2025mmaif}, or self-supervised representation learning \cite{zhao2024equivariant,zhao2023cddfuse}. However, these methods mainly use pretrained knowledge as auxiliary features or side information and do not turn it into direct supervision for fusion. In contrast, our method abandons this long-standing practice and uses external knowledge from powerful frozen pretrained representation models to build a supervision paradigm better aligned with the fundamental objective of MMIF.

\vspace{-2mm}
\section{Methodology}
\begin{figure}
 \centering
  \captionsetup{aboveskip=1pt}
  \includegraphics[width=1\textwidth]{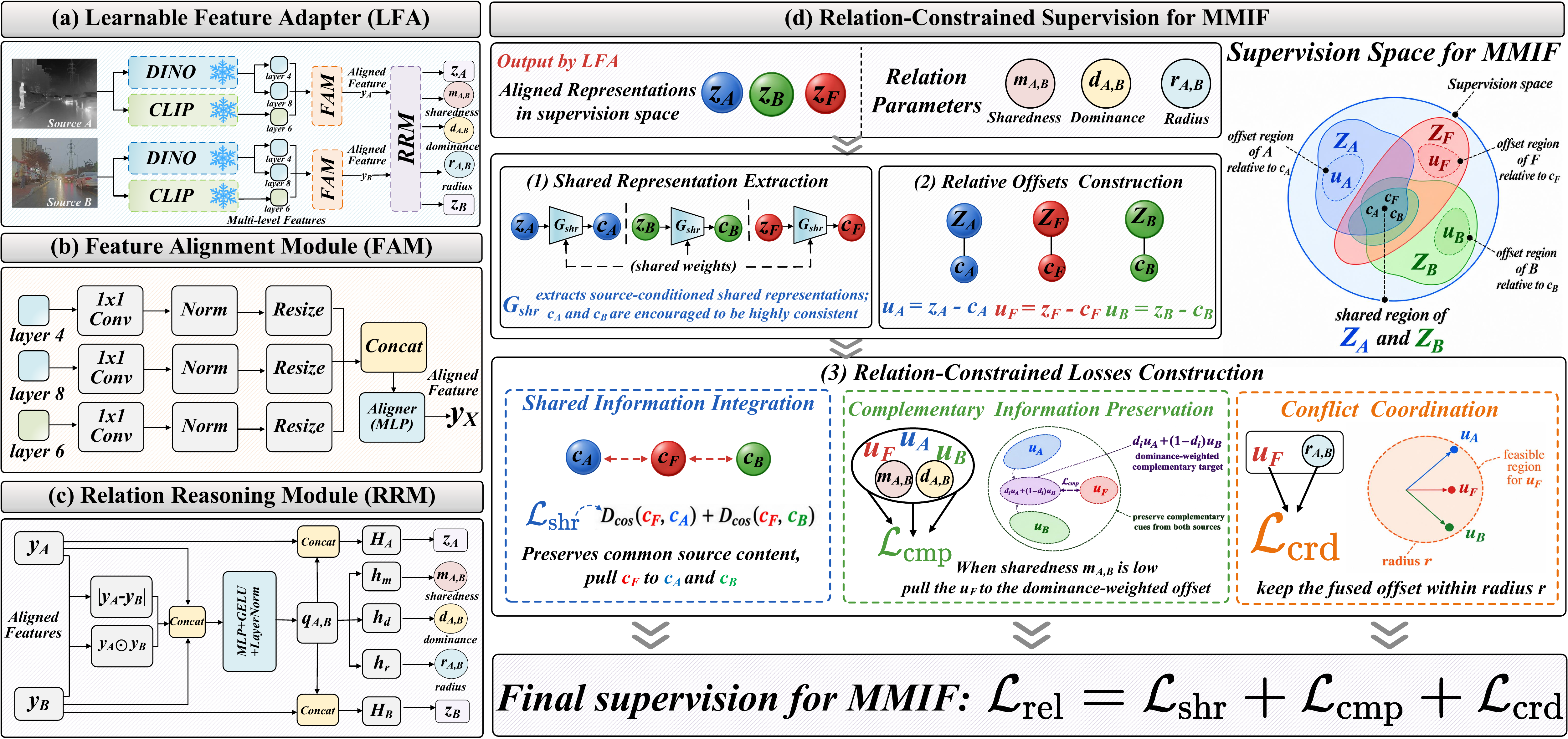}
  \caption{Overview of the Learnable Feature Adapter and Relation-Constrained Supervision.}
  \label{second_figure}
\end{figure}

\textbf{Overview.} As shown in Fig.~\ref{second_figure}, our method consists of three tightly coupled components: a learnable feature adapter (LFA), relation-constrained supervision (RCS) for MMIF, and a tailored training strategy. The LFA maps heterogeneous multi-level features from frozen DINO~\cite{oquab2023dinov2} and CLIP~\cite{CLIP} into a unified supervision space and infers source relations. Based on these outputs, we construct the RCS with three losses for shared information integration, complementary cues preservation, and conflict coordination. Finally, we train the LFA with a self-supervised contrastive learning strategy and optimize it alternately with the fusion net. Appendix~\ref{app:Problem Statement and Modeling} gives the formal problem statement.

\subsection{Learnable Feature Adapter (LFA)}
The learnable feature adapter (LFA) contains two components: a Feature Alignment Module (FAM) and a Relation Reasoning Module (RRM). For each source image $X \in \{A,B\}$, frozen DINO and CLIP extract three feature maps from DINO layer 4, CLIP layer 6, and DINO layer 8, denoted by $f_{D}^{4}(X)$, $f_{C}^{6}(X)$, and $f_{D}^{8}(X)$, respectively. FAM takes these three feature maps as input and outputs an aligned representation $y_X$ in the unified supervision space. RRM then takes the aligned source pair $(y_A,y_B)$ as input and outputs two pair-conditioned source representations, $z_A$ and $z_B$, together with three relation parameters: sharedness $m_{A,B}$, dominance $d_{A,B}$, and coordination radius $r_{A,B}$.

\textbf{Feature Alignment Module.}
To address the mismatch in spatial layout and representation structure across models and feature depths, FAM processes the multi-level features of a source pair $(A,B)$ with a shared projection-and-alignment pipeline. Specifically, for each source image $X \in \{A,B\}$, FAM projects the three feature maps to the same channel dimension and a common token grid:
{\setlength{\abovedisplayskip}{3.5pt}%
\setlength{\belowdisplayskip}{3.5pt}%
\begin{equation}
\tilde{f}_{D}^{4}(X)=P_{D}^{4}\!\left(f_{D}^{4}(X)\right), \quad
\tilde{f}_{C}^{6}(X)=P_{C}^{6}\!\left(f_{C}^{6}(X)\right), \quad
\tilde{f}_{D}^{8}(X)=P_{D}^{8}\!\left(f_{D}^{8}(X)\right),
\end{equation}
}where $P_{D}^{4}$, $P_{C}^{6}$, and $P_{D}^{8}$ denote the corresponding projection operators, and $\tilde{f}_{D}^{4}(X)$, $\tilde{f}_{C}^{6}(X)$, and $\tilde{f}_{D}^{8}(X)$ denote the projected features. Each projector is implemented as a $1 \times 1$ convolution followed by normalization and bilinear resizing, so that the three inputs share the same token layout. The projected features are then flattened into token sequences, concatenated along the channel dimension, and fused by a cross-model aligner:
{\setlength{\abovedisplayskip}{3.5pt}%
\setlength{\belowdisplayskip}{3.5pt}%
\begin{equation}
y_{X}=G_{\mathrm{align}}\!\left([\tilde{f}_{D}^{4}(X),\tilde{f}_{C}^{6}(X),\tilde{f}_{D}^{8}(X)]\right), \quad X \in \{A,B\},
\end{equation}
}where $[\cdot]$ denotes concatenation and $G_{\mathrm{align}}$ denotes a lightweight MLP-based aligner. Applying this shared pipeline to the two source images produces the aligned representations $y_A$ and $y_B$.

\textbf{Relation Reasoning Module.}
Given the aligned source representations $y_A$ and $y_B$ produced by FAM, RRM infers the relation of the source pair in the unified supervision space. To support this relation inference, RRM forms a pair input by concatenating the two aligned source representations, their element-wise absolute difference, and their element-wise product. The absolute difference captures the magnitude of source discrepancy, while the element-wise product reflects their feature-wise agreement. This pair input is first encoded into a shared pair feature $q_{A,B}$ and then projected to two pair-conditioned source representations $z_A$ and $z_B$:
{\setlength{\abovedisplayskip}{3.5pt}%
\setlength{\belowdisplayskip}{3.5pt}%
\begin{equation}
q_{A,B}=E_{\mathrm{pair}}([y_A,y_B,|y_A-y_B|,y_A \odot y_B]), \ \
z_A=H_A([q_{A,B},y_A]), \ \
z_B=H_B([q_{A,B},y_B]),
\end{equation}
}where $E_{\mathrm{pair}}$ denotes the pair encoder, and $H_A$ and $H_B$ denote two source-specific projection heads. In our implementation, $E_{\mathrm{pair}}$ is a lightweight MLP with GELU and LayerNorm, and $H_A$ and $H_B$ are two linear heads followed by normalization.

Based on the shared pair feature $q_{A,B}$, RRM further predicts three relation parameters:
{\setlength{\abovedisplayskip}{3.5pt}%
\setlength{\belowdisplayskip}{3.5pt}%
\begin{equation}
m_{A,B}=\mathrm{Sigmoid}(h_m(q_{A,B})), \quad
d_{A,B}=\mathrm{Sigmoid}(h_d(q_{A,B})), \quad
r_{A,B}=\mathrm{Softplus}(h_r(q_{A,B})),
\end{equation}
}where $h_m$, $h_d$, and $h_r$ are lightweight prediction heads, each implemented by a linear layer, GELU, and LayerNorm. Accordingly, $m_{A,B}$ measures sharedness, $d_{A,B}$ quantifies the tendency toward source $A$ in the non-shared part, with $1-d_{A,B}$ giving the complementary contribution of source $B$, and $r_{A,B}$ defines the coordination radius. Together, the representations $z_A$ and $z_B$ and the relation parameters $m_{A,B}$, $d_{A,B}$, and $r_{A,B}$ are used to construct the subsequent supervision space for MMIF.
\vspace{-2mm}
\subsection{Relation-Constrained Supervision (RCS) for MMIF}
We now use the outputs of LFA to construct relation-constrained supervision for MMIF in the unified supervision space. For a source pair $(A,B)$, RRM provides two pair-conditioned source representations, $z_A$ and $z_B$, together with three relation parameters, $m_{A,B}$, $d_{A,B}$, and $r_{A,B}$. We obtain $z_F$ by encoding the self-pair $(F,F)$ with the same pair-conditioned LFA.

Since the source images contain both shared and non-shared information, a fixed target representation is insufficient for fusion supervision. We introduce a pretrained shared token encoder $G_{\mathrm{shr}}$, which is applied with shared weights to the aligned source representations to extract their shared components:
{\setlength{\abovedisplayskip}{3.5pt}%
\setlength{\belowdisplayskip}{3.5pt}%
\begin{equation}
c_{A}=G_{\mathrm{shr}}(z_{A}), \qquad
c_{B}=G_{\mathrm{shr}}(z_{B}).
\end{equation}
}where $c_{A}$ and $c_{B}$ denote source-conditioned shared representations. During pretraining, $G_{\mathrm{shr}}$ is encouraged to make them highly consistent, so they retain information commonly supported by both modalities and jointly define the consensus reference. The details are in the Appendix~\ref{app:Implementation Details}.

Next, we define the relative offsets of the two source and fused representations as:
{\setlength{\abovedisplayskip}{3.5pt}%
\setlength{\belowdisplayskip}{3.5pt}%
\begin{equation}
u_A=z_A-c_{A}, \quad u_B=z_B-c_{B}, \quad u_F=z_F-c_{F},
\end{equation}
}where $c_{F}=G_{\mathrm{shr}}(z_{F})$ denotes the shared representation extracted from the fused representation. $u_A$ and $u_B$ denote the source offsets, and $u_F$ denotes the fusion offset. These quantities allow us to define three losses for shared information integration, complementary information preservation, and conflict coordination.

\textbf{Shared Information Integration.}
The first requirement of MMIF is to preserve information shared by the two source images. Thus, the fused shared representation should stay close to both shared components rather than drift toward either modality. Accordingly, the shared loss takes the form:
{\setlength{\abovedisplayskip}{3.5pt}%
\setlength{\belowdisplayskip}{3.5pt}%
\begin{equation}
\mathcal{L}_{\mathrm{shr}}
=
\frac{1}{N}\sum_{i=1}^{N}
\left[
\mathcal{D}_{\cos}(c_{F,i},c_{A,i})
+
\mathcal{D}_{\cos}(c_{F,i},c_{B,i})
\right].
\end{equation}
}where $N$ denotes the number of tokens, $i$ indexes the token position, and $\mathcal{D}_{\cos}(\cdot,\cdot)$ denotes the cosine distance. Reducing the loss requires $c_{F,i}$ to align with the shared representations $c_{A,i}$ and $c_{B,i}$, which preserves common source content without introducing modality bias.

\textbf{Complementary Information Preservation.}
Fusion should not collapse all information to the shared cues. In regions with low sharedness, the fused result should preserve useful non-shared cues from the two sources. The complementary loss is given by:
{\setlength{\abovedisplayskip}{3.5pt}%
\setlength{\belowdisplayskip}{3.5pt}%
\begin{equation}
\mathcal{L}_{\mathrm{cmp}}=\frac{1}{N}\sum_{i=1}^{N}\left(1-m_i\right)\,
\mathcal{D}_{\cos}\!\left(u_{F,i},\, d_i u_{A,i}+\left(1-d_i\right)u_{B,i}\right),
\end{equation}
}where $d_i u_{A,i}+\left(1-d_i\right)u_{B,i}$ denotes the dominance-weighted complementary target offset. The factor $1-m_i$ makes this term focus on non-shared regions. When the loss decreases, the fused offset is encouraged to follow the dominance-aware direction defined by the two source offsets, which helps preserve complementary information from both modalities.

\textbf{Conflict Coordination.}
Preserving complementary information alone is insufficient under strong cross-modal conflict, because the fused representation may still be driven excessively toward one source. To prevent this, we introduce a coordination constraint that keeps the fused offset within an adaptive range. The coordination loss is defined as:
{\setlength{\abovedisplayskip}{3.5pt}%
\setlength{\belowdisplayskip}{3.5pt}%
\begin{equation}
\mathcal{L}_{\mathrm{crd}}=\frac{1}{N}\sum_{i=1}^{N}\max\!\left(0,\|u_{F,i}\|_2-r_i\right),
\end{equation}
}where $r_i$ denotes the coordination radius at token $i$. This hinge form imposes no penalty when the fused offset stays within the allowed radius. It only penalizes excessive deviation, thereby limiting conflict-driven bias while preserving flexibility for useful complementary cues.

The final relation-constrained supervision is defined as:
{\setlength{\abovedisplayskip}{3.5pt}%
\setlength{\belowdisplayskip}{3.5pt}%
\begin{equation}
\mathcal{L}_{\mathrm{rel}}=
\mathcal{L}_{\mathrm{shr}}
+\mathcal{L}_{\mathrm{cmp}}
+\mathcal{L}_{\mathrm{crd}}.
\end{equation}
}In this way, fusion supervision is no longer defined by direct approximation to the source images in the spatial domain. Instead, it is defined by whether the fused representation satisfies the shared, complementary, and coordination requirements induced by source relations in the proposed supervision space. Appendix ~\ref{app:analysis} analyzes the effects of relation parameters on the supervision losses.

\subsection{Training Strategy}
\begin{figure}
 \centering
  \captionsetup{aboveskip=1pt}
  \includegraphics[width=1\textwidth]{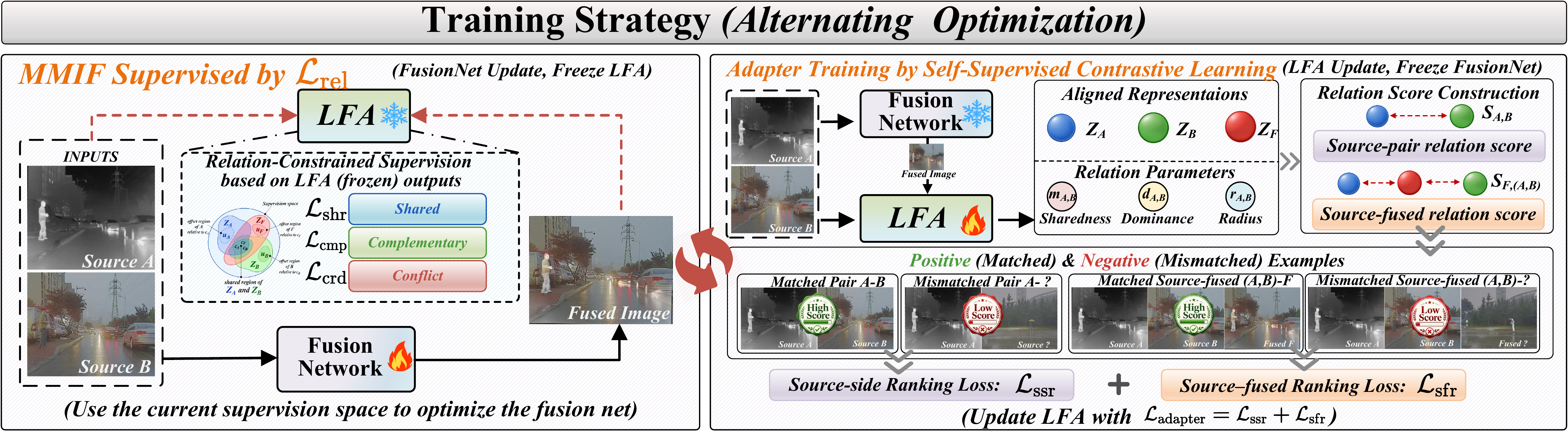}
  \caption{Overview of the Alternating Optimization Strategy.}
  \label{training_figure}
\end{figure}
\textbf{Self-Supervised Contrastive Learning for the Adapter.}
Before using LFA to supervise the fusion network, we pre-train the adapter with a self-supervised contrastive ranking objective to make the learned supervision space discriminative for source relations. Since LFA outputs both pair-conditioned representations and relation parameters, the score used for training should depend on both parts. For a source pair $(A,B)$, we define the source-pair relatedness score as:
{\setlength{\abovedisplayskip}{3.5pt}%
\setlength{\belowdisplayskip}{3.5pt}%
\begin{equation}
s_{A,B}=
\frac{\sum_{i=1}^{N} w_i \cos(z_{A,i},z_{B,i})}
{\sum_{i=1}^{N} w_i+\epsilon},
\quad
w_i=\sigma\!\left(\lambda_m m_i+\lambda_d |2d_i-1|-\lambda_r r_i\right),
\end{equation}
}where $N$ denotes the number of tokens, $\epsilon$ is a small constant, $\cos(\cdot,\cdot)$ denotes cosine similarity, and $\lambda_m$, $\lambda_d$, and $\lambda_r$ are weighting coefficients. In this score, $\cos(z_{A,i},z_{B,i})$ measures the local agreement between the two pair-conditioned representations, while $w_i$ determines how much token $i$ contributes to the final score. The design of $w_i$ follows the roles of the three relation parameters: a larger $m_i$ gives higher weight to shared regions, a larger $|2d_i-1|$ gives higher weight to regions with clearer dominance, and a larger $r_i$ reduces the weight of regions that require a wider coordination range. As a result, $s_{A,B}$ becomes high only when $z_A$ and $z_B$ are locally consistent at tokens where $m_i$, $d_i$, and $r_i$ indicate reliable source relations. For negative source pairs such as $(A,B^-)$ and $(A^-,B)$, the score is computed in the same way after replacing the corresponding source. Appendix~\ref{app:self_supervised_adapter} further explains the relatedness scores and ranking objective.

We first use source-side ranking to make the relation space discriminative at the source-pair level. Let $(A,B^-)$ and $(A^-,B)$ denote two negative pairs formed by replacing one source with an unpaired sample. The source-side ranking loss is:
{\setlength{\abovedisplayskip}{3.5pt}%
\setlength{\belowdisplayskip}{3.5pt}%
\begin{equation}
\mathcal{L}_{\mathrm{ssr}}=
\frac{1}{2}\Big(
\max(0,\delta_{\mathrm{src}}-s_{A,B}+s_{A,B^-})
+
\max(0,\delta_{\mathrm{src}}-s_{A,B}+s_{A^-,B})
\Big),
\end{equation}
}where $\delta_{\mathrm{src}}$ denotes the source-side margin. This margin requires the matched pair to score higher than each mismatched pair by a non-trivial gap, instead of only being slightly larger.

Source-side ranking alone is insufficient, because LFA is ultimately used to supervise fused images. Without source--fused ranking, the learned relation space may not remain reliable when fused images are introduced. We therefore define the relatedness between the fused image and its source pair as:
{\setlength{\abovedisplayskip}{3.5pt}%
\setlength{\belowdisplayskip}{3.5pt}%
\begin{equation}
s_{F,(A,B)}=
\frac{
\sum_{i=1}^{N} w_i
\Big(
m_i c_{shr}
+
(1-m_i)\cos\!\big(u_{F,i},\, d_i u_{A,i}+(1-d_i)u_{B,i}\big)
\Big)
}{
\sum_{i=1}^{N} w_i+\epsilon
},
\end{equation}
}where $c_{shr}\!=\!\cos(c_{F,i},c_{A,i})+\cos(c_{F,i},c_{B,i})$, $u_{A,i}$, $u_{B,i}$, and $u_{F,i}$ are defined in the previous subsection. The first term checks whether the fused representation is close to the shared information, while the second term checks whether it follows the dominance-aware complementary offset. Based on this score, the source--fused ranking loss is:
{\setlength{\abovedisplayskip}{3.5pt}%
\setlength{\belowdisplayskip}{3.5pt}%
\begin{equation}
\mathcal{L}_{\mathrm{sfr}}=
\frac{1}{2}
\Big(
\max(0,\delta_{\mathrm{fuse}}-s_{F,(A,B)}+s_{F,(A,B^{-})})
+
\max(0,\delta_{\mathrm{fuse}}-s_{F,(A,B)}+s_{F,(A^{-},B)})
\Big),
\end{equation}
}where $\delta_{\mathrm{fuse}}$ denotes the fused-side margin, analogous to $\delta_{\mathrm{src}}$ in source-side ranking. The overall adapter objective is: {\setlength{\abovedisplayskip}{3.5pt}%
\setlength{\belowdisplayskip}{3.5pt}%
\begin{equation}
\mathcal{L}_{\mathrm{adapter}}=
\mathcal{L}_{\mathrm{ssr}}+\mathcal{L}_{\mathrm{sfr}}.
\end{equation}
}

\textbf{Alternating optimization.}
Before alternating optimization, the fusion network is first warmed up with a conventional spatial-domain fusion loss to obtain a reasonable baseline performance. Specifically, we adopt the commonly used intensity and gradient constraints:
{\setlength{\abovedisplayskip}{3.5pt}%
\setlength{\belowdisplayskip}{3.5pt}%
\begin{equation}
\mathcal{L}_{\mathrm{warm}}
=
\alpha_1\lVert F-\max(A,B)\rVert_1
+
\alpha_2\lVert\nabla F-\max(\nabla A,\nabla B)\rVert_1,
\end{equation}
}where $\nabla$ denotes the Sobel gradient operator and $\lVert \cdot \rVert_1$ denotes the $L_1$ norm, while $\alpha_1$ and $\alpha_2$ are hyperparameters. This loss is used only for fusion-network warm-up training and is removed once alternating optimization begins.

Next, we switch from conventional fusion loss to the proposed relation-constrained training scheme. The adapter constructs the supervision space, while the fusion network produces the fused images on which this supervision is applied. If either side is kept fixed throughout training, the learned supervision space and the fusion results may remain mismatched. We therefore optimize them alternately. During the adapter-update stage, the fusion network is frozen, and the current fused image $F$ is used to optimize LFA with $\mathcal{L}_{\mathrm{adapter}}$. During the fusion-update stage, LFA is frozen, and the fusion network is optimized with $\mathcal{L}_{\mathrm{rel}}$. This alternating process allows the adapter to refine the supervision space with current fusion results, while the fusion network progressively adapts to the updated supervision. Appendix~\ref{app:pseudocode} provides the full training procedure.

\begin{figure}[t]
 \centering
  \includegraphics[width=1\textwidth]{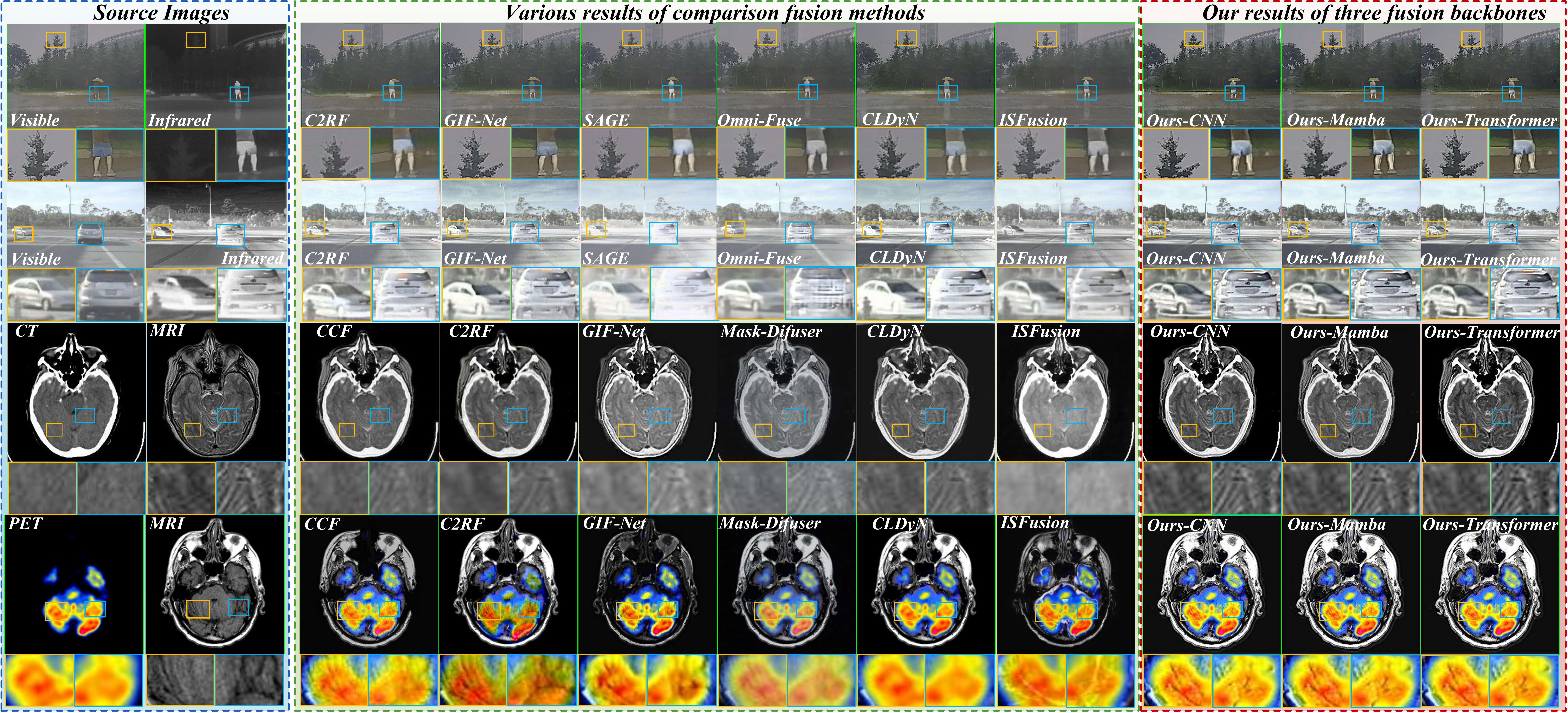}
  \caption{Qualitative comparison for IVF (top two rows) and MIF (bottom two rows).}
  \label{fig:qualitative}
\end{figure}

\begin{table}[t]
\centering
\caption{Quantitative comparison for IVF and MIF task. The best, second best, and third best are highlighted in \best{red}, \second{blue}, and \third{green}, respectively. “Ours(CNN/Mamba/Transformer)” denotes CNN, Mamba, and Transformer fusion backbones trained under our relation-constrained supervision.}
\label{tab:quantitative}
\scriptsize
\setlength{\tabcolsep}{3pt}
\renewcommand{\arraystretch}{0.90}
\resizebox{1\textwidth}{!}{
\begin{tabular}{cc|ccccccc|ccccccc|ccccccc}  
\toprule
\multicolumn{2}{c|}{IVF Task} & \multicolumn{7}{c|}{RoadScene} & \multicolumn{7}{c|}{M$^3$FD} & \multicolumn{7}{c}{MSRS} \\
\midrule
Methods & Pub./Year 
& $AG \uparrow$ & $EN \uparrow$ & $SD \uparrow$ & $MI \uparrow$ & $Q_{AB/F} \uparrow$ & $Q_M \uparrow$ & $Q_P \uparrow$
& $AG \uparrow$ & $EN \uparrow$ & $SD \uparrow$ & $MI \uparrow$ & $Q_{AB/F} \uparrow$ & $Q_M \uparrow$ & $Q_P \uparrow$
& $AG \uparrow$ & $EN \uparrow$ & $SD \uparrow$ & $MI \uparrow$ & $Q_{AB/F} \uparrow$ & $Q_M \uparrow$ & $Q_P \uparrow$ \\
\midrule
EMMA & CVPR 24 & 4.779 & 49.127 & 46.777 & 2.809 & 0.436 & 0.432 & 0.355 & \best{5.871} & 57.569 & 42.377 & 3.771 & 0.589 & 0.512 & 0.460 & 3.788 & 40.539 & 40.590 & 4.164 & 0.541 & 0.760 & 0.496 \\
Text-DiFuse & NIPS 24 & 3.962 & 42.863 & 48.966 & 1.961 & 0.140 & 0.288 & 0.052 & 1.861 & 20.580 & 40.568 & 2.054 & 0.066 & 0.240 & 0.332 & 2.792 & 30.723 & 39.679 & 1.310 & 0.118 & 0.288 & 0.038 \\
Tc-MoA & CVPR 24 & 4.192 & 44.883 & 42.610 & 1.043 & 0.413 & 0.556 & 0.403 & 5.180 & 54.156 & 39.030 & 3.333 & 0.651 & 0.891 & 0.513 & 3.618 & 38.414 & 42.335 & 3.493 & 0.630 & 1.090 & 0.483 \\
ReFusion & IJCV 24 & 5.357 & 59.336 & 48.229 & 2.726 & 0.437 & 0.566 & 0.397 & 5.338 & 57.331 & 40.803 & 3.548 & 0.637 & 0.731 & 0.489 & 3.813 & 40.641 & 42.890 & 4.102 & 0.442 & 1.085 & \best{0.576} \\
C2RF & IJCV 25 & 4.989 & 52.270 & 45.576 & 2.720 & 0.519 & 0.565 & 0.388 & 4.779 & 50.329 & 39.348 & 2.498 & 0.223 & 0.287 & 0.399 & 3.474 & 37.153 & 41.348 & 1.729 & 0.185 & 0.330 & 0.092 \\
GIF-Net & CVPR 25 & \third{5.705} & 59.855 & 45.000 & 2.236 & 0.352 & 0.365 & 0.252 & 5.435 & 52.096 & \third{42.714} & 2.529 & 0.471 & 0.319 & 0.319 & 3.379 & 36.267 & 32.910 & 1.926 & 0.367 & 0.396 & 0.277 \\
SAGE & CVPR 25 & 3.715 & 39.057 & 46.384 & 1.035 & 0.362 & 0.451 & 0.332 & 4.885 & 50.959 & 40.930 & 2.851 & 0.592 & 0.599 & 0.434 & 3.342 & 35.157 & 37.485 & 2.990 & 0.539 & 0.714 & 0.435 \\
Omni-Fuse & TPAMI 25 & 3.206 & 35.212 & 44.830 & 2.187 & 0.303 & 0.396 & 0.267 & 4.514 & 49.067 & 41.029 & 3.208 & 0.430 & 0.362 & 0.277 & 3.091 & 34.123 & 40.043 & 2.664 & 0.361 & 0.412 & 0.217 \\
CLDyN & CVPR 26 & 4.918 & 51.612 & 40.644 & 1.827 & 0.417 & 0.475 & 0.329 & \second{5.824} & 57.312 & 35.254 & 3.235 & \best{0.689} & 0.756 & 0.520 & 3.568 & 37.882 & 39.377 & 2.111 & \third{0.661} & 0.867 & 0.484 \\
ISFusion & CVPR 26 & 5.320 & 54.393 & 35.946 & 2.345 & 0.522 & 0.484 & 0.306 & 5.080 & 52.927 & 36.718 & 2.188 & 0.511 & 0.422 & 0.343 & \best{4.124} & \best{42.850} & 34.100 & 3.303 & 0.491 & 0.550 & 0.357 \\
\midrule
\rowcolor{oursgray}
Ours(CNN) & - - & 5.701 & \third{59.876} & \third{49.673} & \third{2.962} & \third{0.541} & \third{0.604} & \third{0.411} & 5.478 & \third{57.921} & 42.315 & \third{4.238} & 0.664 & \third{1.041} & \third{0.529} & 3.842 & 40.982 & \third{43.517} & \third{4.301} & 0.651 & \third{1.337} & 0.509 \\
\rowcolor{oursgray}
Ours(Mamba) & - - & \second{5.858} & \second{60.742} & \second{50.214} & \second{3.084} & \second{0.552} & \second{0.621} & \second{0.417} & 5.602 & \second{58.736} & \second{42.982} & \second{4.612} & \third{0.671} & \second{1.102} & \second{0.533} & \third{3.902} & \third{41.427} & \second{43.918} & \second{4.493} & \second{0.664} & \second{1.462} & \third{0.518} \\
\rowcolor{oursgray}
Ours(Transformer) & - - & \best{5.861} & \best{61.069} & \best{50.581} & \best{3.179} & \best{0.564} & \best{0.643} & \best{0.425} & \third{5.666} & \best{59.158} & \best{43.407} & \best{4.837} & \second{0.679} & \best{1.157} & \best{0.540} & \second{3.953} & \second{41.854} & \best{44.231} & \best{4.651} & \best{0.678} & \best{1.576} & \second{0.526} \\
\midrule
\multicolumn{2}{c|}{MIF Task} & \multicolumn{7}{c|}{CT-MRI} & \multicolumn{7}{c|}{MRI-PET} & \multicolumn{7}{c}{MRI-SPECT} \\
\midrule
Methods & Pub./Year 
& $AG \uparrow$ & $EN \uparrow$ & $SD \uparrow$ & $MI \uparrow$ & $Q_{AB/F} \uparrow$ & $Q_M \uparrow$ & $Q_P \uparrow$
& $AG \uparrow$ & $EN \uparrow$ & $SD \uparrow$ & $MI \uparrow$ & $Q_{AB/F} \uparrow$ & $Q_M \uparrow$ & $Q_P \uparrow$
& $AG \uparrow$ & $EN \uparrow$ & $SD \uparrow$ & $MI \uparrow$ & $Q_{AB/F} \uparrow$ & $Q_M \uparrow$ & $Q_P \uparrow$ \\
\midrule
EMMA & CVPR 24 & 7.317 & 76.268 & 72.599 & 3.279 & 0.551 & 0.153 & 0.316 & 7.116 & 75.114 & 76.489 & 2.448 & 0.541 & 0.143 & 0.318 & 4.689 & 48.655 & 62.720 & 2.361 & 0.516 & 0.223 & 0.337 \\
Text-DiFuse & NIPS 24 & 6.831 & 70.390 & 66.701 & 3.159 & 0.499 & 0.155 & 0.264 & 8.581 & 88.178 & 66.353 & 2.331 & 0.489 & 0.144 & 0.312 & 5.935 & 60.168 & 58.726 & 2.225 & 0.539 & 0.226 & 0.365 \\
Tc-MoA & CVPR 24 & 6.888 & 72.785 & 72.894 & 3.317 & 0.592 & 0.162 & 0.369 & 7.148 & 75.676 & 69.960 & 2.450 & \second{0.640} & 0.195 & 0.469 & 4.251 & 44.348 & 55.776 & 2.333 & 0.426 & 0.272 & 0.356 \\
ReFusion & IJCV 24 & 8.731 & 88.753 & 76.113 & 3.176 & 0.538 & 0.223 & 0.356 & 7.880 & 80.686 & 72.422 & 2.486 & 0.539 & 0.212 & 0.458 & 5.490 & 55.098 & 61.274 & 2.310 & 0.580 & \best{0.557} & 0.372 \\
C2RF & IJCV 25 & 7.978 & 81.643 & \third{76.192} & 2.953 & 0.332 & 0.171 & 0.131 & 7.933 & 81.310 & 70.626 & 2.183 & 0.458 & 0.216 & 0.284 & 5.369 & 54.366 & 59.497 & 2.257 & 0.547 & 0.400 & 0.309 \\
GIF-Net & CVPR 25 & 9.544 & 94.839 & 70.216 & 3.067 & 0.466 & 0.128 & 0.236 & 6.489 & 67.343 & 61.177 & 2.299 & 0.348 & 0.105 & 0.245 & 4.742 & 48.710 & 51.468 & 2.244 & 0.432 & 0.176 & 0.304 \\
CCF & NIPS 24 & 6.724 & 67.686 & 70.568 & 3.284 & 0.546 & 0.163 & 0.283 & 6.461 & 65.927 & 65.437 & 2.493 & 0.512 & 0.193 & 0.273 & 4.398 & 44.729 & 64.837 & 2.241 & 0.526 & 0.285 & 0.391 \\
Mask-Difuser & TPAMI 25 & 5.984 & 60.253 & 69.600 & 3.299 & 0.484 & 0.166 & 0.213 & 6.126 & 61.783 & 62.930 & \best{2.706} & 0.572 & 0.200 & 0.440 & 4.164 & 41.390 & 58.787 & 2.127 & 0.548 & 0.303 & 0.302 \\
CLDyN & CVPR 26 & 7.777 & 80.208 & 73.716 & 3.245 & 0.597 & 0.206 & 0.394 & 7.618 & 79.791 & 72.779 & 2.466 & \best{0.670} & \third{0.220} & \best{0.490} & 4.919 & 50.583 & 63.594 & \best{2.870} & 0.528 & 0.330 & \best{0.609} \\
ISFusion & CVPR 26 & 8.944 & 88.753 & 72.945 & 2.953 & 0.556 & 0.171 & 0.315 & 8.322 & 81.579 & 66.199 & 2.171 & 0.579 & 0.177 & 0.349 & 5.531 & 54.280 & 58.691 & 2.169 & 0.570 & 0.267 & 0.393 \\
\midrule
\rowcolor{oursgray}
Ours(CNN) & - - & \third{9.558} & \third{94.876} & 75.642 & \third{3.352} & \third{0.611} & \third{0.249} & \third{0.397} & \third{9.127} & \third{96.512} & \third{76.938} & 2.514 & 0.604 & 0.219 & 0.463 & \third{6.145} & \third{63.876} & \third{66.873} & 2.344 & \third{0.581} & 0.423 & 0.396 \\
\rowcolor{oursgray}
Ours(Mamba) & - - & \second{9.573} & \second{95.214} & \second{76.308} & \second{3.421} & \second{0.626} & \second{0.267} & \best{0.443} & \second{9.412} & \second{97.864} & \second{77.614} & \third{2.592} & 0.619 & \second{0.228} & \third{0.474} & \second{6.382} & \second{65.214} & \second{67.942} & \third{2.401} & \second{0.594} & \third{0.451} & \third{0.405} \\
\rowcolor{oursgray}
Ours(Transformer) & - - & \best{9.608} & \best{95.842} & \best{76.527} & \best{3.548} & \best{0.637} & \best{0.290} & \second{0.411} & \best{9.690} & \best{99.031} & \best{78.072} & \second{2.667} & \third{0.627} & \best{0.235} & \second{0.481} & \best{6.571} & \best{66.683} & \best{68.496} & \second{2.448} & \best{0.608} & \second{0.464} & \second{0.413} \\
\bottomrule
\end{tabular}
}
\end{table}

\section{Experiments}
\textbf{Datasets and Backbones.} We evaluate our method on IVF and MIF tasks.
For IVF, we train on 1,000 pairs MSRS~\cite{MSRS} and test on  held-out MSRS (100 pairs), M$^3$FD~\cite{tardal} (100 pairs), and RoadScene~\cite{RoadScene} (50 pairs).
For MIF, we use the Harvard Medical Dataset~\cite{textdifuse} for CT-MRI, MRI-PET, and MRI-SPECT fusion.
\textbf{Since our supervision paradigm is architecture-independent, we adopt CNN-~\cite{zhang2020ifcnn}, Mamba-~\cite{liu2024vmamba}, and Transformer-based~\cite{zamir2022restormer} fusion backbones and provide their details in the Appendix~\ref{app:Implementation Details}.}

\textbf{Training Details.} For IVF, we train on MSRS with 1000 pairs. For MIF, we use 200 training pairs on Harvard Medical Dataset. For each task, the fusion net is first warmed up for 20 epochs with the conventional fusion loss. The adapter is then pretrained for 50 epochs with the self-supervised ranking objective. After initialization, alternating optimization is performed for 5 rounds. In each round, the adapter is updated for 30 epochs with the fusion network frozen, and the
 fusion network is updated for 20 epochs with the adapter frozen. During the fusion-update stage, only the proposed relation-constrained supervision is used. All these training stages use AdamW with an initial learning rate of $3\times10^{-4}$, batch size 4, and no weight decay. The learning rate is halved every 10 epochs. All experiments are conducted on an NVIDIA RTX 4090 GPU. Parameter settings are in the Appendix~\ref{app:Implementation Details}.

\textbf{Comparison methods and metrics.} We compare with EMMA~\cite{zhao2024equivariant}, Text-DiFuse~\cite{textdifuse}, Tc-MoA~\cite{tcmoa}, ReFusion~\cite{bai2025refusion}, C2RF~\cite{c2rf}, GIF-Net~\cite{gifnet}, SAGE~\cite{sage}, Omni-Fuse~\cite{zhang2025omnifuse}, CCF~\cite{ccf}, Mask-Difuser~\cite{maskdifuser}, CLDyN~\cite{CLDyN}, and ISFusion~\cite{ISFusion}, using $AG$~\cite{AG}, $EN$~\cite{EN}, $SD$~\cite{SD}, $MI$~\cite{MI}, $Q_{AB/F}$~\cite{QABF}, $Q_M$~\cite{QM}, and $Q_P$~\cite{QP} as metrics. 

\subsection{Comparison with SOTA}
\textbf{Quantitative Comparison.}
Table~\ref{tab:quantitative} compares our method with recent SOTA methods on IVF and MIF tasks. “Ours(CNN/Mamba/Transformer)” denotes different fusion backbones trained under our relation-constrained supervision. All three variants achieve competitive or superior results across datasets, with the Transformer variant obtaining the best overall performance. These results indicate that relation-space supervision provides a more task-aligned optimization target than spatial-domain source approximation. \textbf{Qualitative Comparison.} Fig.~\ref{fig:qualitative} shows that existing methods often suffer from modality bias, either weakening infrared targets or visible textures in IVF, and either suppressing functional uptake or distorting anatomical boundaries in MIF. In contrast, our method preserves salient targets, structural details, and functional regions more faithfully, producing clearer and more balanced fusion results. Full qualitative comparisons are provided in Appendix~\ref{app:full compare}.

\begin{figure*}[t]
\centering
\captionsetup{aboveskip=4pt, belowskip=2pt}

\begin{minipage}[t]{0.58\textwidth}
    \vspace{0pt}
    \centering
    \includegraphics[width=\linewidth]{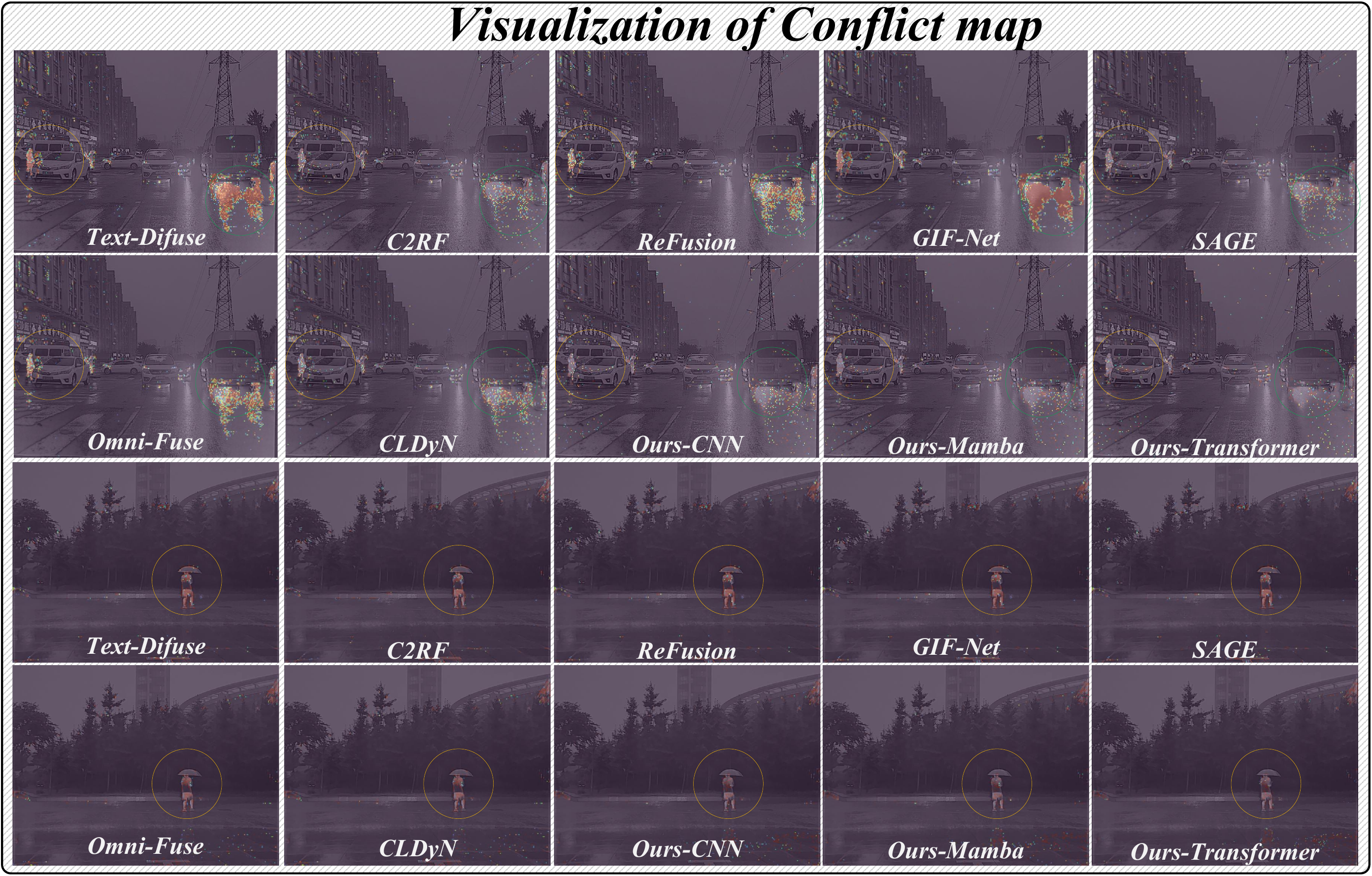}
    \caption{Visualization of conflict map. Warmer colors denote higher source preservation update conflicts, while cooler denote lower conflicts.}
    \label{fig:conflict_map}
\end{minipage}
\hfill
\begin{minipage}[t]{0.39\textwidth}
    \vspace{0pt}
    \centering
    \captionof{table}{Quantitative results on source preservation update conflicts in high-disagreement regions on the M$^3$FD dataset. CR, CS, and UI denote conflict ratio, conflict strength, and update imbalance, respectively.}
    \label{tab:source_conflict_analysis}
    \scriptsize
    \setlength{\tabcolsep}{2pt}
    \renewcommand{\arraystretch}{0.78}
    \resizebox{\linewidth}{!}{%
    \begin{tabular}{l|ccc}
    \toprule
    Method & CR (\%) $\downarrow$ & CS $\downarrow$ & UI $\downarrow$ \\
    \midrule
    Text-DiFuse 
    & 27.25 & 0.806 & 0.4527 \\
    C2RF 
    & 16.04 & 0.715 & 0.3379 \\
    ReFusion 
    & 20.91 & 0.784 & 0.3865 \\
    GIF-Net
    & 18.36 & 0.692 & 0.3514 \\
    SAGE
    & 15.28 & 0.641 & 0.3196 \\
    Omni-Fuse
    & 17.63 & 0.667 & 0.3421 \\
    CLDyN
    & 13.72 & 0.584 & 0.2968 \\
    \midrule
    \rowcolor{oursgray}
    Ours (CNN)
    & \third{7.42} & \third{0.315} & \third{0.218} \\
    \rowcolor{oursgray}
    Ours (Mamba)
    & \second{6.37} & \second{0.287} & \second{0.198} \\
    \rowcolor{oursgray}
    Ours (Transformer)
    & \best{4.77} & \best{0.265} & \best{0.176} \\
    \bottomrule
    \end{tabular}%
    }
\end{minipage}
\end{figure*}

\subsection{Analysis of Source-Preservation Update Conflicts}
Following task-gradient conflict analysis in multi-task optimization~\cite{yu2020gradient,liu2021conflict,navon2022multi,senushkin2023independent}, we view infrared and visible source preservation as two MMIF analysis tasks. This analysis is used only for evaluation. For a fused result $F^m$, we define $\mathcal{L}_{s}^{sp}=\|F^m-I_s\|_1+\lambda_g(\|D_xF^m-D_xI_s\|_1+\|D_yF^m-D_yI_s\|_1)$, where $s\in\{ir,vis\}$ and $D_x,D_y$ are Sobel operators. The induced update is $u_s^m=-\partial\mathcal{L}_{s}^{sp}/\partial F^m$, and patch-level agreement is measured by $\rho_p^m=\cos(u_{ir,p}^m,u_{vis,p}^m)$. A negative $\rho_p^m$ indicates opposite local updates from the two sources. Specifically, we evaluate conflicts on the high-disagreement patch set $\Omega$, which is selected only from source-image intensity and gradient discrepancies and is shared by all methods. On $\Omega$, we report the conflict ratio $\mathrm{CR}=|\Omega|^{-1}\sum_{p\in\Omega}\mathbb{I}(\rho_p^m<0)$ as a percentage, conflict strength $\mathrm{CS}=|\Omega|^{-1}\sum_{p\in\Omega}[-\rho_p^m]_+$, and update imbalance $\mathrm{UI}=|\Omega|^{-1}\sum_{p\in\Omega}|\|u_{ir,p}^{m}\|_2-\|u_{vis,p}^{m}\|_2|/(\|u_{ir,p}^{m}\|_2+\|u_{vis,p}^{m}\|_2+\epsilon)$. Lower values indicate fewer conflicts and better source-side update balance. Details of $\Omega$ and the metrics see Appendix~\ref{app:conflict_detail}. As shown in Fig.~\ref{fig:conflict_map} and Table~\ref{tab:source_conflict_analysis}, existing methods produce stronger high-conflict responses in high-disagreement regions. Ours reduce CR, CS, and UI overall. These results suggest that RCS coordinates cross-modal fusion more stably, and preserves useful source cues with fewer contradictory gradients.

\subsection{Ablation Study}
We conduct ablation studies on IVF and MIF datasets to verify the effectiveness of our main designs. Key results are shown in Table~\ref{tab:ablation_ivf_mif}, and more variants are provided in the Appendix~\ref{app:ablation}.

\textbf{(a) Effect of LFA.} LFA constructs the supervision space from frozen DINO and CLIP features. We evaluate its role through four variants: (1) removing LFA, (2) removing FAM, (3) removing RRM, and (4) replacing RCS with conventional source preservation. Results in Table~\ref{tab:ablation_ivf_mif} (a) show clear performance drops in all cases, especially when LFA or RCS is removed. This shows that pretrained features need task-oriented alignment and relation reasoning for effective fusion supervision.

\textbf{(b) Effect of RCS.}
RCS defines fusion supervision through shared integration, complementary preservation, and conflict coordination. We examine its role by removing $\mathcal{L}_{\mathrm{shr}}$, $\mathcal{L}_{\mathrm{cmp}}$, and $\mathcal{L}_{\mathrm{crd}}$, respectively. Results in Table~\ref{tab:ablation_ivf_mif} (b) show consistent performance drops in all cases. This shows that adaptive relation constraints are essential for task-aligned fusion supervision.

\textbf{(c) Effect of Adapter Training.}
Adapter training makes the learned supervision space discriminative and compatible with fused images. We evaluate its role through three variants: (1) removing $\mathcal{L}_{\mathrm{ssr}}$, (2) removing $\mathcal{L}_{\mathrm{sfr}}$, and (3) fixing the adapter after pretraining. Results in Table~\ref{tab:ablation_ivf_mif} (c) show clear degradation across these variants. This confirms that the adapter should be pretrained with relation ranking and alternately refined with the fusion network.

\begin{table*}[t]
\centering
\captionsetup{belowskip=2pt}
\caption{Ablation study of the proposed method on the IVF RoadScene and MIF MRI-PET datasets. The best results are highlighted in \best{bold}. More ablation variants are provided in the appendix.}
\label{tab:ablation_ivf_mif}
\scriptsize
\setlength{\tabcolsep}{2.1pt}
\renewcommand{\arraystretch}{0.90}
\resizebox{\textwidth}{!}{%
\begin{tabular}{c|c|c|ccccccc|ccccccc}
\toprule
\multirow{2}{*}{Group} 
& \multirow{2}{*}{ID}
& \multirow{2}{*}{Description} 
& \multicolumn{7}{c|}{IVF on RoadScene} 
& \multicolumn{7}{c}{MIF on MRI-PET} \\
\cmidrule(lr){4-10}\cmidrule(lr){11-17}
& 
& 
& $AG \uparrow$ & $EN \uparrow$ & $SD \uparrow$ & $MI \uparrow$ & $Q_{AB/F} \uparrow$ & $Q_M \uparrow$ & $Q_P \uparrow$
& $AG \uparrow$ & $EN \uparrow$ & $SD \uparrow$ & $MI \uparrow$ & $Q_{AB/F} \uparrow$ & $Q_M \uparrow$ & $Q_P \uparrow$ \\
\midrule
\multirow{4}{*}{(a)}
& (1) & w/o LFA
& 5.247 & 58.932 & 47.286 & 2.861 & 0.506 & 0.589 & 0.372
& 8.721 & 94.863 & 72.418 & 2.321 & 0.552 & 0.196 & 0.414 \\
& (2) & w/o FAM
& 5.516 & 60.217 & 49.103 & 3.041 & 0.537 & 0.618 & 0.401
& 9.184 & 97.326 & 75.861 & 2.512 & 0.594 & 0.218 & 0.452 \\
& (3) & w/o RRM
& 5.438 & 59.846 & 48.624 & 2.982 & 0.529 & 0.611 & 0.394
& 9.062 & 96.814 & 75.104 & 2.463 & 0.586 & 0.213 & 0.444 \\
& (4) & w/o RCS
& 5.196 & 58.714 & 47.635 & 2.803 & 0.497 & 0.581 & 0.363
& 8.604 & 94.217 & 71.936 & 2.276 & 0.541 & 0.189 & 0.405 \\
\midrule
\multirow{3}{*}{(b)}
& (1) & w/o $\mathcal{L}_{\mathrm{shr}}$
& 5.642 & 60.384 & 49.427 & 3.086 & 0.545 & 0.626 & 0.410
& 9.346 & 97.942 & 76.412 & 2.574 & 0.607 & 0.224 & 0.461 \\
& (2) & w/o $\mathcal{L}_{\mathrm{cmp}}$
& 5.471 & 60.092 & 48.891 & 2.947 & 0.523 & 0.613 & 0.392
& 9.118 & 97.156 & 75.386 & 2.438 & 0.581 & 0.216 & 0.441 \\
& (3) & w/o $\mathcal{L}_{\mathrm{crd}}$
& 5.573 & 60.263 & 49.018 & 3.018 & 0.532 & 0.607 & 0.386
& 9.241 & 97.624 & 75.927 & 2.496 & 0.592 & 0.211 & 0.433 \\
\midrule
\multirow{3}{*}{(c)}
& (1) & w/o $\mathcal{L}_{\mathrm{ssr}}$
& 5.324 & 59.463 & 48.102 & 2.902 & 0.514 & 0.598 & 0.379
& 8.891 & 95.842 & 73.864 & 2.384 & 0.567 & 0.204 & 0.425 \\
& (2) & w/o $\mathcal{L}_{\mathrm{sfr}}$
& 5.386 & 59.721 & 48.406 & 2.918 & 0.517 & 0.602 & 0.381
& 8.976 & 96.214 & 74.293 & 2.406 & 0.573 & 0.207 & 0.429 \\
& (3) & fixed adapter
& 5.452 & 59.984 & 48.713 & 2.971 & 0.526 & 0.609 & 0.389
& 9.087 & 96.936 & 75.021 & 2.451 & 0.583 & 0.212 & 0.438 \\
\midrule
  \rowcolor{oursgray} \multicolumn{3}{c|}{Full Model (Transformer)}
& \best{5.861} & \best{61.069} & \best{50.581} & \best{3.179} & \best{0.564} & \best{0.643} & \best{0.425}
& \best{9.690} & \best{99.031} & \best{78.072} & \best{2.667} & \best{0.627} & \best{0.235} & \best{0.481} \\
\bottomrule
\end{tabular}%
}
\vspace{-8pt}
\end{table*}

\begin{figure}[t]
 \centering
  \includegraphics[width=1\textwidth]{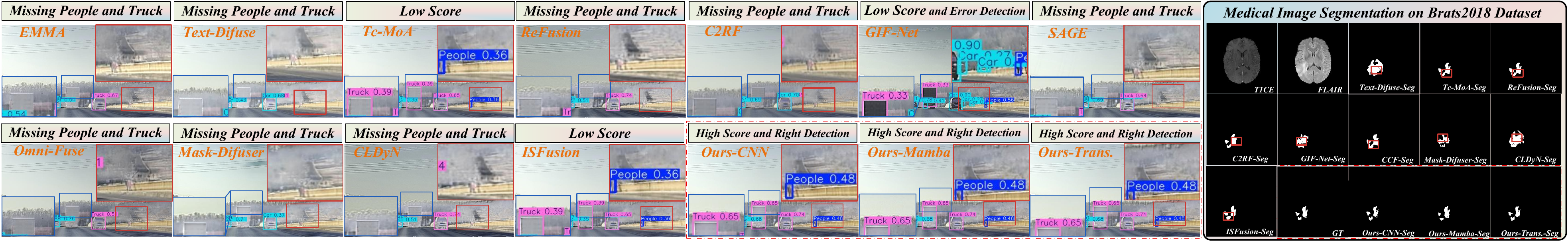}
  \caption{Qualitative results on downstream tasks (Object detection and medical image segmentation).}
  \label{fig:downstream}
\end{figure}

\begin{table}[t]
\centering
\begin{minipage}[t]{0.48\textwidth}
\centering
\caption{Quantitative results of various methods for object detection task on the M$^3$FD~\cite{tardal}.}
\resizebox{\linewidth}{!}{
\begin{tabular}{c|cccccc|c}
\toprule
Object Detection & People & Car & Bus & Motorcycle & Lamp & Truck & mAP \\
\midrule
Infrared & 0.304 & 0.422 & 0.024 & 0.549 & 0.000 & 0.112 & 0.235 \\
Visible & 0.515 & 0.603 & 0.691 & 0.455 & 0.585 & 0.491 & 0.557 \\
EMMA & 0.450 & 0.617 & 0.457 & 0.469 & 0.465 & 0.517 & 0.496 \\
Text-DiFuse & 0.547 & 0.614 & 0.781 & 0.565 & 0.632 & 0.545 & 0.614 \\
Tc-MoA & 0.550 & 0.687 & 0.757 & 0.459 & 0.565 & 0.317 & 0.556 \\
ReFusion & 0.489 & 0.663 & 0.522 & 0.538 & 0.597 & 0.663 & 0.579 \\
GIF-Net & \second{0.695} & 0.782 & 0.756 & 0.528 & 0.565 & 0.477 & 0.634 \\
SAGE & 0.511 & 0.760 & 0.671 & 0.485 & 0.542 & 0.475 & 0.574 \\
Omni-Fuse & 0.614 & 0.739 & \third{0.795} & 0.511 & 0.614 & 0.543 & 0.636 \\
Mask-Difuser & 0.628 & 0.756 & \second{0.802} & 0.536 & 0.641 & 0.579 & 0.657 \\
CLDyN & 0.646 & 0.773 & 0.764 & 0.552 & 0.658 & 0.621 & 0.669 \\
ISFusion & 0.665 & \third{0.784} & 0.736 & \third{0.574} & \second{0.672} & 0.684 & \third{0.686} \\
\midrule
  \rowcolor{oursgray} 
Ours (CNN) & 0.677 & \best{0.813} & 0.709 & 0.427 & 0.655 & \second{0.753} & 0.672 \\
  \rowcolor{oursgray} 
Ours (Mamba) & \third{0.683} & \second{0.791} & \best{0.823} & \second{0.591} & \best{0.689} & \third{0.712} & \second{0.715} \\
  \rowcolor{oursgray} 
Ours (Transformer) & \best{0.701} & 0.781 & 0.774 & \best{0.633} & \third{0.669} & \best{0.822} & \best{0.730} \\
\bottomrule
\end{tabular}}
\label{tab:OD}
\end{minipage}
\hfill
\begin{minipage}[t]{0.48\textwidth}
\centering
\caption{Quantitative results of various methods for medical segmentation task on the BraTS2018~\cite{Brats2018}.}
\resizebox{\linewidth}{!}{
\begin{tabular}{c|ccccc}   
\toprule
\textbf{Case} & T1CE & FLAIR & EMMA & Text-DiFuse & Tc-MoA \\
\midrule
\textbf{IoU~\cite{IoU}}  & 0.695 & 0.657 & 0.743 & 0.685 & 0.689 \\
\textbf{Dice~\cite{Dice}} & 0.807 & 0.770 & 0.841 & 0.751 & 0.829 \\
\midrule
\textbf{Case} & ReFusion & C2RF & GIF-Net & CCF & Mask-Difuser \\
\midrule
\textbf{IoU}  & 0.639 & 0.724 & 0.639 & 0.661 & 0.723 \\
\textbf{Dice} & 0.757 & 0.833 & 0.758 & 0.552 & 0.829 \\
\midrule
\textbf{Case} & CLDyN & ISFusion & Ours (CNN) & Ours (Mamba) & Ours (Transformer) \\
\midrule
\textbf{IoU}  & 0.745 & 0.694 & \third{0.782} & \second{0.791} & \best{0.799} \\
\textbf{Dice} & 0.843 & 0.806 & \third{0.846} & \best{0.854} & \second{0.850} \\
\bottomrule
\end{tabular}
}
\label{tab:MS}
\end{minipage}
\end{table}
\subsection{Downstream Tasks}
We further evaluate downstream perception using fused images. For IVF, all methods are tested with the same pretrained YOLOv12~\cite{tian2025yolov12} detector without fine-tuning; for MIF, medical fusion results are evaluated with UniverSeg~\cite{butoi2023universeg}. As shown in Fig.~\ref{fig:downstream} and Tables~\ref{tab:OD}--\ref{tab:MS}, our method achieves more reliable detection and segmentation results than competing methods. These gains indicate that relation-constrained supervision better preserves task-relevant cues and improves downstream utility.

\section{Conclusion}
In this paper, we present a relation-constrained supervision paradigm that moves MMIF supervision beyond the spatial domain. This paradigm learns a relation space from frozen pretrained representations and aligns supervision with the goal of fusion. Our adapter aligns heterogeneous DINO and CLIP features and infers relation parameters for shared integration, complementary preservation, and conflict coordination. With adapter-tailored contrastive ranking and alternating training, the proposed space is effectively optimized. Extensive experiments validate its effectiveness.

\section*{Acknowledgments}
This work was supported by the National Natural Science Foundation of China [No.62401097, 62601484]; Fundamental Research Funds for Central Universities, Dalian Minzu University [No.0854-53]; Liaoning Province Applied Basic Research Program [2026JH2/101300163]; Liaoning Province Science and Technology Joint Plan (2024JH2/102600113).

\section*{Competing Interests} The authors declare no competing interests.

{
    \small
    \bibliographystyle{abbrv}
    \bibliography{nips.bib} 
}


\newpage
\appendix
\section{Pseudocode of the Training Procedure}
\label{app:pseudocode}
We provide the pseudocode of the proposed training procedure in Algorithm~\ref{alg:training_procedure} and Algorithm~\ref{alg:adapter_objective}. Algorithm~\ref{alg:training_procedure} summarizes the full optimization pipeline. It first warms up the fusion network with a conventional fusion loss to obtain initial fused images. This warm-up is only used to provide preliminary fused samples for adapter learning. After that, the fusion network is fixed and the adapter is initialized with the self-supervised ranking objective. The later training follows an alternating scheme. In each round, LFA is first updated for several epochs using the current fused images, and the fusion network is then updated for several epochs using only the relation-constrained supervision. The conventional fusion loss is not used in this alternating stage.

\begin{algorithm}[h]
\caption{Training procedure of the proposed framework}
\label{alg:training_procedure}
\begin{algorithmic}[1]
\Statex \textbf{Input:} Paired source dataset $\mathcal{D}=\{(A,B)\}$.
\Statex \textbf{Modules:} Frozen feature extractor $\Psi$, fusion network $\mathcal{F}_{\Theta}$, learnable feature adapter $\mathcal{A}_{\Phi}$.
\State Initialize parameters $\Theta$ and $\Phi$.

\Statex \textit{// Warm-up fusion network to obtain initial fused images}
\For{epoch $=1$ to $E_{\mathrm{warm}}$}
    \For{each mini-batch $(A,B)\subset\mathcal{D}$}
        \State $F=\mathcal{F}_{\Theta}(A,B)$
        \State Update $\Theta$ by back-propagating $\mathcal{L}_{\mathrm{warm}}(F;A,B)$
    \EndFor
\EndFor

\Statex \textit{// Initialize adapter with fixed fusion network}
\State Freeze $\Theta$.
\For{epoch $=1$ to $E_{\mathrm{ada}}$}
    \For{each mini-batch $(A,B)\subset\mathcal{D}$}
        \State Sample unpaired sources $A^{-}$ and $B^{-}$.
        \State $F=\mathcal{F}_{\Theta}(A,B)$
        \State $\mathcal{L}_{\mathrm{adapter}}=\textsc{AdapterObjective}(A,B,A^{-},B^{-},F;\Phi)$
        \State Update $\Phi$ by back-propagating $\mathcal{L}_{\mathrm{adapter}}$
    \EndFor
\EndFor

\Statex \textit{// Alternating optimization}
\For{round $=1$ to $K$}

    \Statex \textit{// Adapter update: train LFA for several epochs with current fused images}
    \State Freeze $\Theta$ and unfreeze $\Phi$.
    \For{epoch $=1$ to $E_{\mathrm{ada}}^{\mathrm{alt}}$}
        \For{each mini-batch $(A,B)\subset\mathcal{D}$}
            \State Sample unpaired sources $A^{-}$ and $B^{-}$.
            \State $F=\mathcal{F}_{\Theta}(A,B)$
            \State $\mathcal{L}_{\mathrm{adapter}}=\textsc{AdapterObjective}(A,B,A^{-},B^{-},F;\Phi)$
            \State Update $\Phi$ by back-propagating $\mathcal{L}_{\mathrm{adapter}}$
        \EndFor
    \EndFor

    \Statex \textit{// Fusion update: train FusionNet for several epochs only with RCS}
    \State Freeze $\Phi$ and unfreeze $\Theta$.
    \For{epoch $=1$ to $E_{\mathrm{fus}}^{\mathrm{alt}}$}
        \For{each mini-batch $(A,B)\subset\mathcal{D}$}
            \State $F=\mathcal{F}_{\Theta}(A,B)$
            \State $\mathcal{L}_{\mathrm{rel}}=\mathrm{RCS}(A,B,F;\Phi)$
            \State Update $\Theta$ by back-propagating $\mathcal{L}_{\mathrm{rel}}$
        \EndFor
    \EndFor

\EndFor

\Statex \textbf{Output:} Trained fusion network $\mathcal{F}_{\Theta}$ and adapter $\mathcal{A}_{\Phi}$.
\end{algorithmic}
\end{algorithm}

Algorithm~\ref{alg:adapter_objective} details the adapter objective used in both adapter initialization and adapter-update stages. Given a matched source pair, two unpaired sources, and the current fused image, the adapter first computes source-pair scores for the matched and mismatched pairs. These scores are used to form the source--source ranking loss $\mathcal{L}_{\mathrm{ssr}}$. The fused image is then encoded by feeding the self-pair $(F,F)$ into the same pair-conditioned adapter, producing the fused representation $z_F$. Based on $z_F$, the adapter computes one matched source--fused score $s_{F,(A,B)}$ and two mismatched source--fused scores $s_{F,(A,B^-)}$ and $s_{F,(A^-,B)}$. These scores are used to form the source--fused ranking loss $\mathcal{L}_{\mathrm{sfr}}$, which encourages the fused image to be more compatible with its matched source pair than with mismatched source pairs. The final adapter objective is the sum of these two ranking losses. This design trains the adapter to distinguish matched source relations and to remain reliable when fused images enter the supervision space.

\begin{algorithm}[h]
\caption{\textsc{AdapterObjective}$(A,B,A^{-},B^{-},F;\Phi)$}
\label{alg:adapter_objective}
\begin{algorithmic}[1]
\Statex \textbf{Input:} Matched source pair $(A,B)$, unpaired sources $A^{-}$ and $B^{-}$, fused image $F$.
\Statex \textbf{Module:} Learnable feature adapter $\mathcal{A}_{\Phi}$.
\Statex \textbf{Margins:} $\delta_{\mathrm{src}}$, $\delta_{\mathrm{fuse}}$.

\State Extract $\Psi(X)=\{f_D^4(X),f_C^6(X),f_D^8(X)\}$ for $X\in\{A,B,A^{-},B^{-},F\}$.

\Statex \textit{// Source--source ranking}
\State $\mathcal{A}_{\Phi}(\Psi(A),\Psi(B))
\rightarrow (z_A^{A,B},z_B^{A,B},m_{A,B},d_{A,B},r_{A,B}) \rightarrow s_{A,B}$.
\State $\mathcal{A}_{\Phi}(\Psi(A),\Psi(B^{-}))
\rightarrow (z_A^{A,B^-},z_{B^-}^{A,B^-},m_{A,B^-},d_{A,B^-},r_{A,B^-}) \rightarrow s_{A,B^{-}}$.
\State $\mathcal{A}_{\Phi}(\Psi(A^{-}),\Psi(B))
\rightarrow (z_{A^-}^{A^-,B},z_B^{A^-,B},m_{A^-,B},d_{A^-,B},r_{A^-,B}) \rightarrow s_{A^{-},B}$.

\State Compute source--source ranking loss:
{\setlength{\abovedisplayskip}{3.5pt}%
\setlength{\belowdisplayskip}{3.5pt}%
\begin{equation}
\mathcal{L}_{\mathrm{ssr}}=
\frac{1}{2}
\Big(
\max(0,\delta_{\mathrm{src}}-s_{A,B}+s_{A,B^-})
+
\max(0,\delta_{\mathrm{src}}-s_{A,B}+s_{A^-,B})
\Big).
\end{equation}
}
\Statex \textit{// Matched source pair should score higher than mismatched source pairs}

\Statex \textit{// Source--fused ranking}
\State Encode the fused self-pair:
$\mathcal{A}_{\Phi}(\Psi(F),\Psi(F))
\rightarrow z_F$.
\State $(z_A^{A,B},z_B^{A,B},m_{A,B},d_{A,B},r_{A,B},z_F)
\rightarrow s_{F,(A,B)}$.
\State $(z_A^{A,B^-},z_{B^-}^{A,B^-},m_{A,B^-},d_{A,B^-},r_{A,B^-},z_F)
\rightarrow s_{F,(A,B^-)}$.
\State $(z_{A^-}^{A^-,B},z_B^{A^-,B},m_{A^-,B},d_{A^-,B},r_{A^-,B},z_F)
\rightarrow s_{F,(A^-,B)}$.

\State Compute source--fused ranking loss:
{\setlength{\abovedisplayskip}{3.5pt}%
\setlength{\belowdisplayskip}{3.5pt}%
\begin{equation}
\mathcal{L}_{\mathrm{sfr}}=
\frac{1}{2}
\Big(
\max(0,\delta_{\mathrm{fuse}}-s_{F,(A,B)}+s_{F,(A,B^-)})
+
\max(0,\delta_{\mathrm{fuse}}-s_{F,(A,B)}+s_{F,(A^-,B)})
\Big).
\end{equation}
}
\Statex \textit{// Fused image should score higher with its matched source pair than with mismatched source pairs}

\State Compute adapter objective:
{\setlength{\abovedisplayskip}{3.5pt}%
\setlength{\belowdisplayskip}{3.5pt}%
\begin{equation}
\mathcal{L}_{\mathrm{adapter}}=
\mathcal{L}_{\mathrm{ssr}}+\mathcal{L}_{\mathrm{sfr}}.
\end{equation}
}

\Statex \textbf{Output:} $\mathcal{L}_{\mathrm{adapter}}$.
\end{algorithmic}
\end{algorithm}

\section{Problem Statement and Modeling}
\label{app:Problem Statement and Modeling}
Given two source images $A$ and $B$, the goal of MMIF is to learn a fusion network $\mathcal{F}$ that generates a fused image: $F = \mathcal{F}(A,B;\Theta)$, where $\Theta$ denotes the parameters of the fusion network. Under this notation, most existing multimodal image fusion methods follow a spatial source-image supervision paradigm, which directly uses the source images as GT to compute loss, defined as:
{
\setlength{\abovedisplayskip}{3.5pt}%
\setlength{\belowdisplayskip}{3.5pt}%
\begin{equation}
\arg\min_{\Theta}
\mathcal{L}_{\mathrm{spa}}(F;A,B).
\end{equation}
}Here, $\mathcal{L}_{\mathrm{spa}}$ denotes the losses composed of spatial-domain criteria such as intensity, gradient, and structural similarity. However, $A$ and $B$ are carriers of information, rather than natural ground truth for the fused image. They typically contain shared information, modality-specific information, and potential conflicts simultaneously. Therefore, directly using spatial-domain source images as GT often drives fusion learning toward pixel-level compromise or modality bias, rather than faithfully modeling the integration objective of MMIF. To address this limitation, we argue that an ideal supervisory space for MMIF should align with the fundamental objective of fusion. Since MMIF aims to integrate shared information, preserve complementary information, and handle potential cross-modal conflicts, we construct the proposed supervisory space from these three aspects. In other words, the proposed fusion supervision is defined not by direct spatial-domain approximation to source images, but by whether the fused result satisfies the fundamental requirements of MMIF in a knowledge-driven relational space. Under this formulation, fusion supervision is defined as:
{
\setlength{\abovedisplayskip}{3.5pt}%
\setlength{\belowdisplayskip}{3.5pt}%
\begin{equation}
\arg\min_{\Theta}\;
\!\!\!\!\!\!\underbrace{\mathcal{L}_{\mathrm{rel}}}_{\mathcal{L}_{\mathrm{shr}},\mathcal{L}_{\mathrm{cmp}},\mathcal{L}_{\mathrm{crd}}}\!\!\!\!\!\!\!\!\!
\Big(
\underbrace{\mathcal{A}\big(\Psi(A),\Psi(B),\Psi(F);\Phi\big)}_{ \hat{A},\hat{B},\hat{F},m,d,r}
\Big).
\end{equation}
}Here, $\Phi$ denotes the parameters of the learnable feature adapter $\mathcal{A}$, which outputs aligned features $z_{A}$, $z_{B}$, and $z_{F}$ in the unified supervision space, along with three relation parameters $m$, $d$, and $r$. Built on these outputs, $\mathcal{L}_{\mathrm{rel}}$ denotes the proposed relation-constrained supervision with three components: the shared loss $\mathcal{L}_{\mathrm{shr}}$, the complementary loss $\mathcal{L}_{\mathrm{cmp}}$, and the coordination loss $\mathcal{L}_{\mathrm{crd}}$.

\section{Why Fusion Supervision Requires Sharedness, Complementarity, and Coordination?}
\label{app:three_relations}

A proper supervision space for MMIF should reflect the objective of fusion rather than the appearance of any single source image. Given a source pair $(A,B)$, the fused image is expected to integrate information that is jointly supported by both sources, retain useful information that appears mainly in one source, and avoid unstable updates caused by cross-modal conflicts. These requirements naturally lead to three relations: sharedness, complementarity, and coordination.

\textbf{Sharedness.}
Sharedness describes the degree to which two sources provide consistent information at a local position. When the two sources contain the same structure or semantic content, the fused representation should not be biased toward either source. Instead, it should preserve their common information. Without modeling sharedness, a supervision loss may treat the two source images as two independent targets and force the fusion model to make a pixel-level compromise. This may weaken common structures or introduce modality bias. Therefore, sharedness is needed to identify where the fused representation should follow the common representation.

\textbf{Complementarity.}
Complementarity describes useful non-shared information provided by different sources. In MMIF, some important cues are visible only in one modality, such as thermal targets in infrared images or fine textures in visible images. A supervision space that only emphasizes shared information would push the fused representation toward common cues and may suppress modality-specific cues. Therefore, complementarity is needed to guide the fused representation to preserve non-shared but useful information. In our formulation, this is achieved by supervising the fused offset rather than only its distance to the shared representations.

\textbf{Coordination.}
Coordination is required because not all source differences are useful complementarity. Some regions contain strong cross-modal disagreement, noise, or inconsistent responses. If the supervision always forces the fused representation to follow one source or preserve both sources without restriction, the fusion result may be driven excessively toward one modality. This can introduce artifacts or unstable representations. Coordination defines an adaptive tolerance range for the fused offset. It allows useful complementary deviations while suppressing excessive deviations caused by conflict.

Together, the three relations define different but connected aspects of fusion supervision. Sharedness characterizes the degree of common information between the sources and modulates the strength of complementary supervision. Complementarity determines how non-shared source cues should be preserved. Coordination determines how far the fused representation can deviate under conflict. Thus, the proposed supervision space is not defined by direct spatial approximation to source images, but by whether the fused representation satisfies these three requirements induced by source relations.

\section{Additional Analysis of the Relation-Constrained Supervision}
\label{app:analysis}

This appendix provides a detailed explanation of how the relation parameters affect the supervision structure, the relation-constrained losses, the relatedness scores, and the training objective. The goal is to clarify how the proposed supervision space responds to different source relations.

\subsection{Effects of Relation Parameters on Supervision}
\textbf{Sharedness and Dominance ($m_i$ and $d_i$).}
The sharedness parameter $m_i$ characterizes the degree of shared information between the two modalities at token $i$ and modulates the complementary loss through the factor $1-m_i$. The dominance parameter $d_i$ controls the relative contribution of the two sources in non-shared regions through the dominance-aware complementary offset:
{\setlength{\abovedisplayskip}{3.5pt}%
\setlength{\belowdisplayskip}{3.5pt}%
\begin{equation}
v_i=d_i u_{A,i}+\left(1-d_i\right)u_{B,i}.
\end{equation}
}Here, $v_i$ denotes the dominance-aware complementary offset. When $d_i$ is close to $1$ or $0$, the target offset favors source $A$ or $B$, respectively, while $d_i$ close to $0.5$ gives both sources more balanced contributions. The complementary loss is defined as:
{\setlength{\abovedisplayskip}{3.5pt}%
\setlength{\belowdisplayskip}{3.5pt}%
\begin{equation}
\mathcal{L}_{\mathrm{cmp}}=\frac{1}{N}\sum_{i=1}^{N}\left(1-m_i\right)
\mathcal{D}_{\cos}(u_{F,i},v_i).
\end{equation}
}Thus, higher sharedness weakens the complementary constraint, while lower sharedness strengthens it to preserve modality-specific cues. Meanwhile, reducing $\mathcal{L}_{\mathrm{cmp}}$ encourages the fused offset $u_{F,i}$ to follow the dominance-aware direction in non-shared regions. The term $|2d_i-1|$ is also used in the relatedness score. It becomes larger when the dominance relation is clearer, allowing such tokens to provide stronger relation evidence during adapter ranking.

\textbf{Coordination radius $r_i$.}
The coordination radius $r_i$ controls the allowed range of the fused offset. Its effect appears in the coordination loss:
{\setlength{\abovedisplayskip}{3.5pt}%
\setlength{\belowdisplayskip}{3.5pt}%
\begin{equation}
\mathcal{L}_{\mathrm{crd}}=\frac{1}{N}\sum_{i=1}^{N}\max\!\left(0,\|u_{F,i}\|_2-r_i\right).
\end{equation}
}When $r_i$ is small, the fused offset is constrained to stay close to the source-induced reference. When $r_i$ is large, the loss becomes more tolerant and allows a wider offset range. This is useful in regions with strong cross-modal conflict, where strict alignment may suppress useful information. However, a large $r_i$ also indicates that the local relation is less stable. Therefore, the relatedness score reduces the contribution of such tokens through the negative term $-\lambda_r r_i$ in the relation-aware weight. In this way, $r_i$ weakens overly strict supervision in difficult regions while preventing unstable tokens from dominating adapter ranking.

\section{Why Self-Supervised Contrastive Learning Works for LFA?}
\label{app:self_supervised_adapter}

LFA must learn pair-conditioned representations and relation parameters before it can supervise the fusion network. Since MMIF has no ground-truth fused image, these outputs cannot be trained with direct target labels. We instead use the pairing structure of source images as a self-supervised signal. For a matched source pair $(A,B)$, the two images are captured from the same scene and share objects, structures, and semantic layout. For mismatched pairs such as $(A,B^-)$ and $(A^-,B)$, this scene-level correspondence is broken. This provides a natural contrastive signal for adapter learning.

\subsection{Source-Side Contrastive Signal}

The source-side signal trains LFA to distinguish matched source pairs from mismatched ones. Instead of requiring an absolute ground-truth relation score, it only requires the matched pair to have a higher relatedness score:
{\setlength{\abovedisplayskip}{3.5pt}%
\setlength{\belowdisplayskip}{3.5pt}%
\begin{equation}
s_{A,B} > s_{A,B^-}, \qquad s_{A,B} > s_{A^-,B}.
\end{equation}
}This relative constraint is suitable for LFA because the adapter is expected to learn a relation space, rather than predict a fixed label for each pair. The corresponding source--source ranking loss is defined as:
{\setlength{\abovedisplayskip}{3.5pt}%
\setlength{\belowdisplayskip}{3.5pt}%
\begin{equation}
\mathcal{L}_{\mathrm{ssr}}=
\frac{1}{2}\Big(
\max(0,\delta_{\mathrm{src}}-s_{A,B}+s_{A,B^-})
+
\max(0,\delta_{\mathrm{src}}-s_{A,B}+s_{A^-,B})
\Big).
\end{equation}
}The margin $\delta_{\mathrm{src}}$ prevents weak separation. Thus, LFA cannot satisfy the objective by assigning similar scores to all pairs. It must learn representations and relation parameters that separate matched and mismatched source relations.

\subsection{Coupled Training of Representations and Relation Parameters}

The relatedness score is designed to depend on both types of LFA outputs. For a source pair $(A,B)$, the score is defined as:
{\setlength{\abovedisplayskip}{3.5pt}%
\setlength{\belowdisplayskip}{3.5pt}%
\begin{equation}
s_{A,B}=
\frac{\sum_{i=1}^{N}w_i\cos(z_{A,i},z_{B,i})}
{\sum_{i=1}^{N}w_i+\epsilon},
\quad
w_i=\sigma\!\left(\lambda_m m_i+\lambda_d |2d_i-1|-\lambda_r r_i\right).
\end{equation}
}
The cosine term trains the pair-conditioned representations $z_A$ and $z_B$. It requires matched source pairs to be locally consistent and mismatched pairs to be less consistent. The weight $w_i$ trains the relation parameters. A larger $m_i$ gives higher weight to shared regions, and a larger $|2d_i-1|$ gives higher weight to regions with clearer dominance. A larger $r_i$ reduces the weight because a wider coordination range indicates a less stable local relation.

This design prevents one part of LFA from being optimized alone. If $z_A$ and $z_B$ are not discriminative, matched and mismatched pairs cannot be separated. If $m_i$, $d_i$, and $r_i$ do not identify reliable relation evidence, unstable tokens may dominate the score. Therefore, the ranking objective jointly calibrates the pair-conditioned representations and the relation parameters through the same score function.

\subsection{Source-Fused Contrastive Signal}

Source-side ranking alone is insufficient because LFA is later used to supervise fused images. If the adapter only sees source-source pairs during training, it may not remain reliable when a fused image enters the supervision space. To reduce this mismatch, we include the current fused image $F$ in adapter training and impose the source--fused ranking constraint:
{\setlength{\abovedisplayskip}{3.5pt}%
\setlength{\belowdisplayskip}{3.5pt}%
\begin{equation}
s_{F,(A,B)} > s_{F,(A^-,B)},\qquad s_{F,(A,B)} > s_{F,(A,B^-)}.
\end{equation}
}Here, $s_{F,(A,B)}$ measures the compatibility between the fused representation and the relation-aware reference induced by its matched source pair, while $s_{F,(A^-,B)}$ and $s_{F,(A,B^-)}$ measure its compatibility with references induced by mismatched source pairs. A valid fused representation should integrate the shared and complementary information from its corresponding sources. Therefore, it should better match its reference from the matched source pair than those from mismatched pairs.


The source--fused score follows the same structure as the relation-constrained supervision:
{\setlength{\abovedisplayskip}{3.5pt}%
\setlength{\belowdisplayskip}{3.5pt}%
\begin{equation}
s_{F,(A,B)}=
\frac{
\sum_{i=1}^{N} w_i
\Big(
m_i c_{shr}
+
(1-m_i)\cos\!\big(u_{F,i},\, d_i u_{A,i}+(1-d_i)u_{B,i}\big)
\Big)
}{
\sum_{i=1}^{N} w_i+\epsilon
}.
\end{equation}
}where $c_{shr}\!=\!\cos(c_{F,i},c_{A,i})+\cos(c_{F,i},c_{B,i})$. The first term checks whether $z_{F,i}$ is close to the shared cues in shared regions. The second term checks whether $u_{F,i}$ follows the dominance-aware complementary offset in non-shared regions. Thus, source--fused ranking does not simply add fused images as extra samples. It trains LFA on the same type of relation that is later used for fusion supervision.

The corresponding source--fused ranking loss is defined as:
{\setlength{\abovedisplayskip}{3.5pt}%
\setlength{\belowdisplayskip}{3.5pt}%
\begin{equation}
\mathcal{L}_{\mathrm{sfr}}=
\frac{1}{2}
\Big(
\max(0,\delta_{\mathrm{fuse}}-s_{F,(A,B)}+s_{F,(A,B^{-})})
+
\max(0,\delta_{\mathrm{fuse}}-s_{F,(A,B)}+s_{F,(A^{-},B)})
\Big).
\end{equation}
}Here, the margin $\delta_{\mathrm{fuse}}$ plays the same role as $\delta_{\mathrm{src}}$, but it is applied to source--fused ranking. The full adapter objective is:
{\setlength{\abovedisplayskip}{3.5pt}%
\setlength{\belowdisplayskip}{3.5pt}%
\begin{equation}
\mathcal{L}_{\mathrm{adapter}}=
\mathcal{L}_{\mathrm{ssr}}+\mathcal{L}_{\mathrm{sfr}}.
\end{equation}
}

\subsection{Sufficient Pretraining and Alternating Optimization}

The adapter receives supervision from both pair discrimination and fused-image consistency. For each matched source pair, the source--source ranking loss compares it with two mismatched pairs. This trains LFA to identify reliable source relations. The source--fused ranking loss then introduces the current fused image into adapter learning. Since both scores depend on $z$ and the relation parameters $m,d,r$, these objectives jointly update the representation space and the relation parameters.

The warm-up fusion network provides the initial fused images required by source--fused ranking. These fused images do not need to be optimal. They only provide a starting distribution for adapting LFA from source-source relations to source-fused relations. After warm-up, LFA is trained with the fusion network fixed. The current fused image $F$ is generated by the fixed fusion network and is used in $\mathcal{L}_{\mathrm{sfr}}$.

The following training proceeds alternately. During the adapter-update stage, the fusion network is frozen, and LFA is updated with $\mathcal{L}_{\mathrm{adapter}}$ using the current fused images. During the fusion-update stage, LFA is frozen, and the fusion network is optimized only with $\mathcal{L}_{\mathrm{rel}}$. No conventional fusion loss is used in this stage. This creates a closed training loop: improved fused images provide more informative source--fused samples for LFA, and an improved LFA provides a more reliable supervision space for the fusion network.

\begin{figure}[t]
 \centering
  \includegraphics[width=1\textwidth]{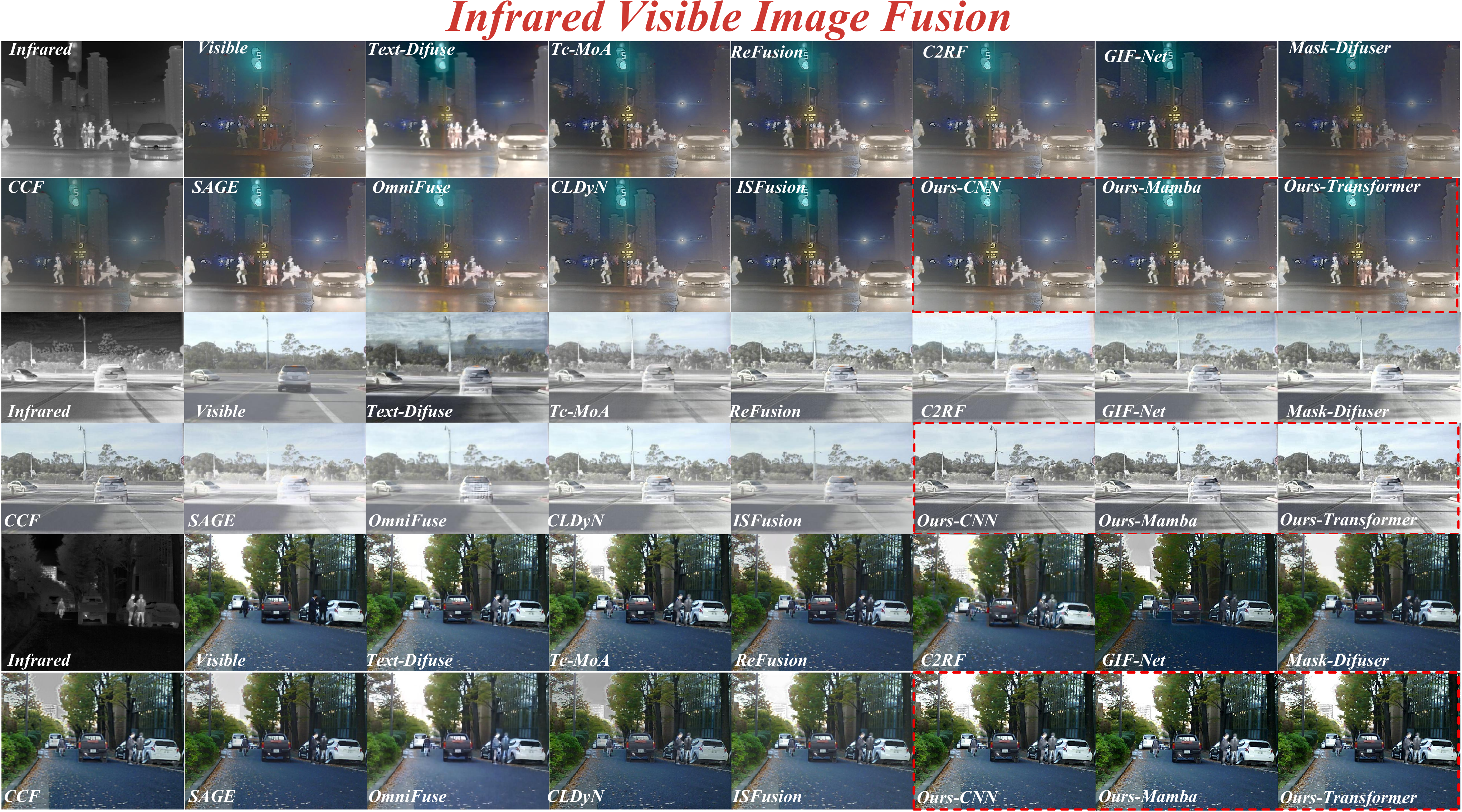}
  \caption{Full qualitative comparison for IVF task on M$^3$FD, RoadScene and MSRS.}
  \label{fig:sup-ivf}
\end{figure}

\section{Full Qualitative Comparisons}
\label{app:full compare}
We provide full qualitative comparisons for both MIF and IVF tasks. For MIF, the results are reported on CT-MRI~\cite{textdifuse}, MRI-PET~\cite{textdifuse}, and MRI-SPECT~\cite{textdifuse} fusion. For IVF, the results are reported on M$^3$FD~\cite{tardal}, RoadScene~\cite{RoadScene}, and MSRS~\cite{MSRS}. Since our supervision paradigm is independent of the fusion architecture, we include CNN-, Mamba-, and Transformer-based variants in all comparisons.

\begin{figure}[h]
 \centering
  \includegraphics[width=1\textwidth]{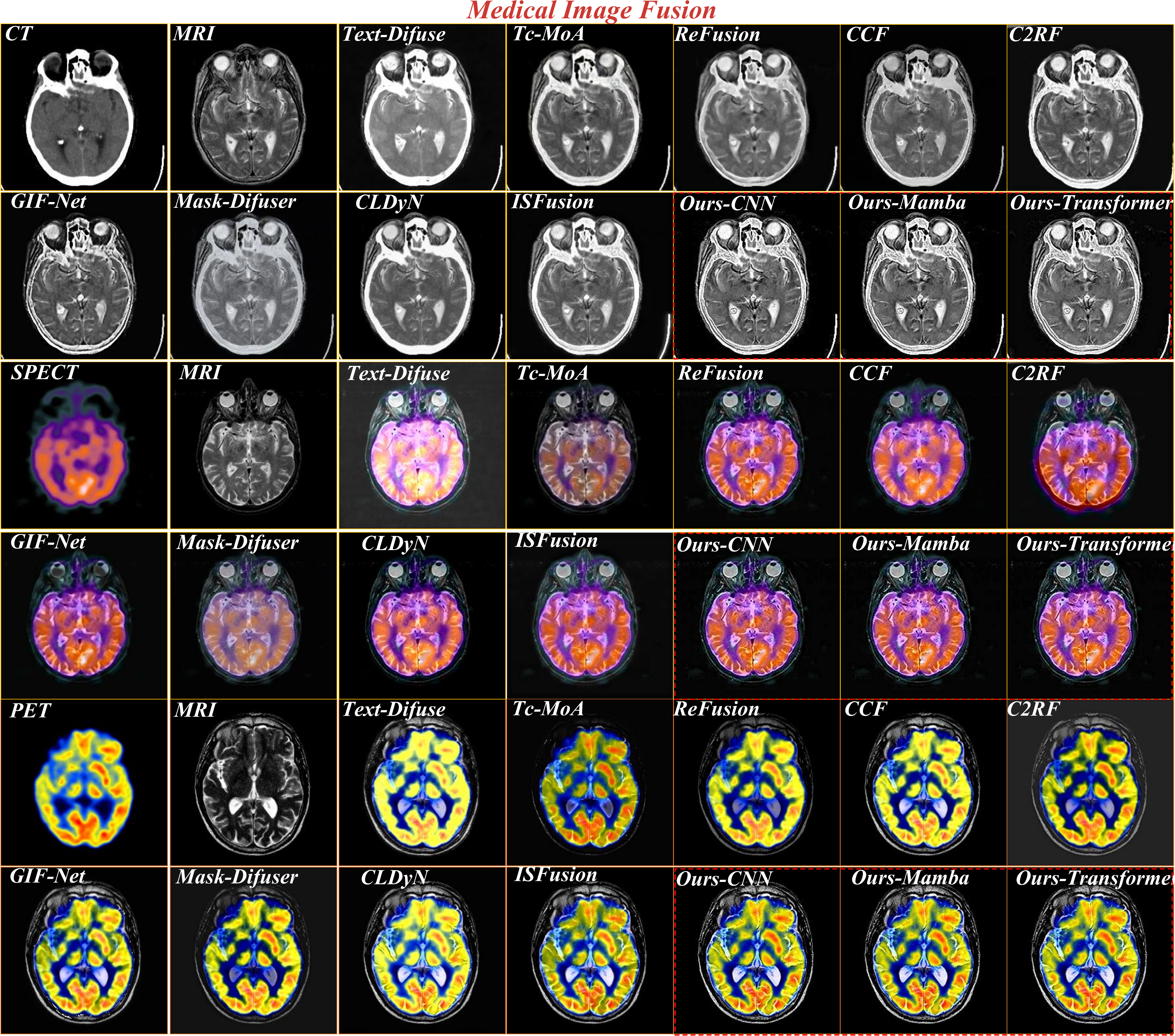}
  \caption{Full qualitative comparison for MIF task on CT-MRI, MRI-PET and MRI-SPECT.}
  \label{fig:sup-mif}
\end{figure}

As shown in Fig.~\ref{fig:sup-mif}, our method produces visually reliable medical fusion results across all three MIF settings. In CT-MRI fusion, our variants preserve clear anatomical structures while maintaining soft-tissue contrast from MRI. In MRI-PET and MRI-SPECT fusion, our results retain functional color information and avoid excessive color bleeding or structural distortion. These results show that the proposed supervision can guide different backbones to preserve complementary medical information in a stable manner.

Fig.~\ref{fig:sup-ivf} further shows the full qualitative comparison on IVF datasets. Existing methods may suffer from weak thermal targets, over-smoothed textures, or insufficient visible-detail preservation in challenging night and road scenes. In contrast, our CNN, Mamba, and Transformer variants consistently preserve salient infrared targets while keeping visible structures, such as vehicles, pedestrians, roads, and background textures. This indicates that the proposed relation-constrained supervision is not tied to a specific backbone and can provide effective guidance across different fusion architectures.

\section{Details of Source-Preservation Update Conflict Analysis}
\label{app:conflict_detail}
We provide the detailed procedure for the source-preservation update conflict analysis used in Table~\ref{tab:source_conflict_analysis}. Following task-gradient conflict analysis in multi-task optimization~\cite{yu2020gradient,liu2021conflict,navon2022multi,senushkin2023independent}, we use the update directions induced by two source-preservation objectives to examine whether a fused result can accommodate infrared and visible information without strong local conflict. This analysis is only used for evaluation and is not used to train any model.

For a fused result $F^m$ produced by method $m$, we define the source-preservation objective for source $s\in\{ir,vis\}$ as:
\[
\mathcal{L}_{s}^{sp}
=
\|F^m-I_s\|_1
+
\lambda_g
\left(
\|D_xF^m-D_xI_s\|_1
+
\|D_yF^m-D_yI_s\|_1
\right),
\]
where $I_s$ denotes the source image, and $D_x,D_y$ are Sobel operators. Here, $\|\cdot\|_1$ denotes the mean absolute error over pixels. The induced update direction is:
\[
u_s^m
=
-\frac{\partial \mathcal{L}_{s}^{sp}}{\partial F^m}.
\]
The negative sign converts the loss gradient into the local descent direction. Since both source-preservation gradients are negated, the cosine-based conflict criterion is unchanged compared with using the original loss gradients.

For each patch $p$, we restrict the two update maps to this patch and flatten them into vectors, denoted as $u_{ir,p}^m$ and $u_{vis,p}^m$. Their directional agreement is:
\[
\rho_p^m
=
\frac{
\langle u_{ir,p}^m,u_{vis,p}^m\rangle
}{
\|u_{ir,p}^m\|_2\|u_{vis,p}^m\|_2+\epsilon
}.
\]
A negative $\rho_p^m$ means that infrared and visible preservation require opposite local updates for the same fused result.

\textbf{High-disagreement patch set $\Omega$.}
We evaluate the conflict on high-disagreement regions, where the two source images are more likely to contain competing information. For each patch $p$, we compute a source discrepancy score:
\[
d_p
=
\frac{1}{|p|}
\sum_{q\in p}
\left(
|I_{ir}^{q}-I_{vis}^{q}|
+
|D_xI_{ir}^{q}-D_xI_{vis}^{q}|
+
|D_yI_{ir}^{q}-D_yI_{vis}^{q}|
\right).
\]
The high-disagreement patch set $\Omega$ is defined as the top 30\% patches ranked by $d_p$. Since $\Omega$ is computed only from the two source images, it is fixed for each image pair and shared by all compared methods.

\textbf{Conflict ratio.}
The conflict ratio is defined as:
\[
\mathrm{CR}
=
\frac{1}{|\Omega|}
\sum_{p\in\Omega}
\mathbb{I}(\rho_p^m<0).
\]
CR measures how frequently opposite source-preservation updates occur. A lower CR means that fewer local regions require conflicting corrections to preserve the two sources.

\textbf{Conflict strength.}
The conflict strength is defined as:
\[
\mathrm{CS}
=
\frac{1}{|\Omega|}
\sum_{p\in\Omega}
[-\rho_p^m]_+.
\]
CS measures how severe the conflicts are. Unlike CR, which only counts whether a conflict exists, CS considers the magnitude of negative alignment. A value close to $1$ indicates nearly opposite update directions, while a value close to $0$ indicates weak or no conflict.

\textbf{Update imbalance.}
The update imbalance is defined as:
\[
\mathrm{UI}
=
\frac{1}{|\Omega|}
\sum_{p\in\Omega}
\frac{
\left|
\|u_{ir,p}^{m}\|_2-\|u_{vis,p}^{m}\|_2
\right|
}{
\|u_{ir,p}^{m}\|_2+\|u_{vis,p}^{m}\|_2+\epsilon
}.
\]
UI is not a cosine-based conflict measure. It measures the normalized magnitude gap between the infrared and visible source-preservation updates. This metric checks whether a low conflict score is caused by suppressing one modality. A lower UI indicates that the two source-preservation updates have more balanced magnitudes.

Together, CR, CS, and UI provide complementary evidence. CR measures how often conflicts occur, CS measures how strong these conflicts are, and UI checks whether conflict reduction is achieved without suppressing one source-preservation update.

\section{Broader Impact of the Proposed Paradigm}
Table~\ref{tab:broader_plug_in} evaluates whether our paradigm can serve as a general refinement strategy for existing fusion models. For a fair plug-in evaluation, we keep the original network architecture, training dataset, and training settings of each baseline unchanged. The only modification is to replace its original supervision with our relation-constrained supervision paradigm. We apply this setting to three representative baselines, including ReFusion, EMMA, and C2RF, and report the results on the RoadScene dataset.

The upgraded versions consistently improve all seven metrics, showing that the gains are not tied to a specific backbone. EMMA* obtains large improvements on $EN$, $Q_M$, and $Q_P$, while ReFusion* and C2RF* also achieve clear gains on structural and perceptual quality metrics. These results indicate that relation-constrained supervision provides a transferable optimization signal. It can enhance source preservation, structural fidelity, and fusion quality across different model designs. Therefore, the proposed paradigm has broader applicability beyond our own architecture and can improve existing image fusion systems with limited modification.

\begin{table}
\centering
\caption{Refinement effects of our paradigm on the RoadScene dataset. Original vs.\ upgraded models (* denotes our paradigm inserted).}
\label{tab:broader_plug_in}
\scriptsize
\setlength{\tabcolsep}{2.1pt}
\renewcommand{\arraystretch}{0.90}
\resizebox{0.88\textwidth}{!}{%
\begin{tabular}{l|ccccccc}
\toprule
Method & $AG\!\uparrow$ & $EN\!\uparrow$ & $SD\!\uparrow$ & $MI\!\uparrow$ & $Q_{AB/F}\!\uparrow$ & $Q_M\!\uparrow$ & $Q_P\!\uparrow$ \\
\midrule
ReFusion~\cite{bai2025refusion}    & 5.357 & 59.336 & 48.229 & 2.726 & 0.437 & 0.566 & 0.397 \\
ReFusion*   & 5.692 & 60.108 & 49.356 & 2.916 & 0.523 & 0.612 & 0.418 \\
\textbf{Gain} & \textcolor{bestred}{6.3\%\,$\uparrow$} & \textcolor{bestred}{1.3\%\,$\uparrow$} & \textcolor{bestred}{2.3\%\,$\uparrow$} & \textcolor{bestred}{7.0\%\,$\uparrow$} & \textcolor{bestred}{19.7\%\,$\uparrow$} & \textcolor{bestred}{8.1\%\,$\uparrow$} & \textcolor{bestred}{5.3\%\,$\uparrow$} \\
\midrule
EMMA~\cite{zhao2024equivariant}        & 4.779 & 49.127 & 46.777 & 2.809 & 0.436 & 0.432 & 0.355 \\
EMMA*       & 5.514 & 58.742 & 49.108 & 2.943 & 0.518 & 0.589 & 0.407 \\
\textbf{Gain} & \textcolor{bestred}{15.4\%\,$\uparrow$} & \textcolor{bestred}{19.6\%\,$\uparrow$} & \textcolor{bestred}{5.0\%\,$\uparrow$} & \textcolor{bestred}{4.8\%\,$\uparrow$} & \textcolor{bestred}{18.8\%\,$\uparrow$} & \textcolor{bestred}{36.3\%\,$\uparrow$} & \textcolor{bestred}{14.6\%\,$\uparrow$} \\
\midrule
C2RF~\cite{c2rf}        & 4.989 & 52.270 & 45.576 & 2.720 & 0.519 & 0.565 & 0.388 \\
C2RF*       & 5.603 & 58.196 & 48.962 & 2.935 & 0.548 & 0.606 & 0.412 \\
\textbf{Gain} & \textcolor{bestred}{12.3\%\,$\uparrow$} & \textcolor{bestred}{11.3\%\,$\uparrow$} & \textcolor{bestred}{7.4\%\,$\uparrow$} & \textcolor{bestred}{7.9\%\,$\uparrow$} & \textcolor{bestred}{5.6\%\,$\uparrow$} & \textcolor{bestred}{7.3\%\,$\uparrow$} & \textcolor{bestred}{6.2\%\,$\uparrow$} \\
\bottomrule
\end{tabular}%
}
\end{table}

\section{Extended Ablation Study}
\label{app:ablation}
We provide the full ablation results in Table~\ref{tab:extended_ablation_ivf_mif}. Besides the key variants reported in the main paper, we further analyze adaptive relation parameters and adapter training choices. These additional results show consistent trends on both IVF and MIF.

\textbf{(a) Effect of adaptive relation parameters.}
We examine whether the relation parameters in RCS should be adaptively inferred through three variants: (1) fixing the sharedness, (2) fixing dominance, and (3) fixing the coordination radius. Results in Table~\ref{tab:extended_ablation_ivf_mif} show that all three variants underperform the full model. Fixed sharedness weakens the distinction between shared and non-shared information. Fixed dominance prevents the model from adapting to source-specific contributions. Fixed radius reduces the flexibility of conflict coordination. These results show that sharedness, dominance, and coordination radius should be inferred from each source pair.

\textbf{(b) Effect of relation-aware scoring.}
We evaluate the token weighting strategy by replacing the relation-aware weight $w_i$ with uniform weights. This variant treats all tokens equally during adapter ranking and consistently degrades performance. The result indicates that not all regions contribute equally to relation learning. Tokens with reliable sharedness, clear dominance, and smaller coordination uncertainty should receive higher weights. This verifies the necessity of relation-aware scoring in the adapter objective.

\textbf{(c) Effect of adapter pretraining.}
We study the role of adapter pretraining by directly starting alternating optimization from an untrained adapter. This setting leads to a clear performance drop on both IVF and MIF. Since the adapter defines the supervision space, an unstable adapter can provide noisy relation constraints to the fusion network. Pretraining gives LFA a discriminative initialization before it is used for fusion supervision.

\textbf{(d) Effect of fixed adapter.}
We further test a fixed-adapter variant, where LFA is pretrained once and then kept frozen during fusion training. This variant performs worse than the full model. The result shows that a static supervision space is insufficient because fused images evolve as the fusion network improves. Updating LFA during alternating optimization allows the supervision space to adapt to current fused results.

\begin{table*}[t]
\centering
\caption{Extended ablation study of the proposed method on the IVF RoadScene and MIF MRI-PET datasets. This table includes additional variants for fixed relation parameters and adapter training strategies. The best results are highlighted in \best{bold}.}
\label{tab:extended_ablation_ivf_mif}
\scriptsize
\setlength{\tabcolsep}{2.1pt}
\renewcommand{\arraystretch}{0.90}
\resizebox{\textwidth}{!}{%
\begin{tabular}{c|c|c|ccccccc|ccccccc}
\toprule
\multirow{2}{*}{Group} 
& \multirow{2}{*}{ID}
& \multirow{2}{*}{Description} 
& \multicolumn{7}{c|}{IVF on RoadScene} 
& \multicolumn{7}{c}{MIF on MRI-PET} \\
\cmidrule(lr){4-10}\cmidrule(lr){11-17}
& 
& 
& $AG \uparrow$ & $EN \uparrow$ & $SD \uparrow$ & $MI \uparrow$ & $Q_{AB/F} \uparrow$ & $Q_M \uparrow$ & $Q_P \uparrow$
& $AG \uparrow$ & $EN \uparrow$ & $SD \uparrow$ & $MI \uparrow$ & $Q_{AB/F} \uparrow$ & $Q_M \uparrow$ & $Q_P \uparrow$ \\
\midrule
\multirow{4}{*}{(a)}
& (1) & w/o LFA
& 5.247 & 58.932 & 47.286 & 2.861 & 0.506 & 0.589 & 0.372
& 8.721 & 94.863 & 72.418 & 2.321 & 0.552 & 0.196 & 0.414 \\
& (2) & w/o FAM
& 5.516 & 60.217 & 49.103 & 3.041 & 0.537 & 0.618 & 0.401
& 9.184 & 97.326 & 75.861 & 2.512 & 0.594 & 0.218 & 0.452 \\
& (3) & w/o RRM
& 5.438 & 59.846 & 48.624 & 2.982 & 0.529 & 0.611 & 0.394
& 9.062 & 96.814 & 75.104 & 2.463 & 0.586 & 0.213 & 0.444 \\
& (4) & w/o RCS
& 5.196 & 58.714 & 47.635 & 2.803 & 0.497 & 0.581 & 0.363
& 8.604 & 94.217 & 71.936 & 2.276 & 0.541 & 0.189 & 0.405 \\
\midrule
\multirow{6}{*}{(b)}
& (1) & w/o $\mathcal{L}_{\mathrm{shr}}$
& 5.642 & 60.384 & 49.427 & 3.086 & 0.545 & 0.626 & 0.410
& 9.346 & 97.942 & 76.412 & 2.574 & 0.607 & 0.224 & 0.461 \\
& (2) & w/o $\mathcal{L}_{\mathrm{cmp}}$
& 5.471 & 60.092 & 48.891 & 2.947 & 0.523 & 0.613 & 0.392
& 9.118 & 97.156 & 75.386 & 2.438 & 0.581 & 0.216 & 0.441 \\
& (3) & w/o $\mathcal{L}_{\mathrm{crd}}$
& 5.573 & 60.263 & 49.018 & 3.018 & 0.532 & 0.607 & 0.386
& 9.241 & 97.624 & 75.927 & 2.496 & 0.592 & 0.211 & 0.433 \\
& (4) & fixed sharedness
& 5.589 & 60.338 & 49.214 & 3.052 & 0.539 & 0.621 & 0.404
& 9.281 & 97.735 & 76.103 & 2.536 & 0.601 & 0.221 & 0.456 \\
& (5) & fixed dominance
& 5.492 & 60.041 & 48.776 & 2.963 & 0.526 & 0.614 & 0.395
& 9.146 & 97.208 & 75.542 & 2.455 & 0.584 & 0.217 & 0.445 \\
& (6) & fixed radius
& 5.601 & 60.171 & 49.006 & 3.011 & 0.533 & 0.609 & 0.388
& 9.265 & 97.531 & 75.818 & 2.487 & 0.590 & 0.212 & 0.436 \\
\midrule
\multirow{6}{*}{(c)}
& (1) & w/o $\mathcal{L}_{\mathrm{ssr}}$
& 5.324 & 59.463 & 48.102 & 2.902 & 0.514 & 0.598 & 0.379
& 8.891 & 95.842 & 73.864 & 2.384 & 0.567 & 0.204 & 0.425 \\
& (2) & w/o $\mathcal{L}_{\mathrm{sfr}}$
& 5.386 & 59.721 & 48.406 & 2.918 & 0.517 & 0.602 & 0.381
& 8.976 & 96.214 & 74.293 & 2.406 & 0.573 & 0.207 & 0.429 \\
& (3) & uniform $w_i$
& 5.538 & 60.064 & 48.957 & 3.006 & 0.531 & 0.612 & 0.393
& 9.203 & 97.382 & 75.674 & 2.481 & 0.589 & 0.215 & 0.442 \\
& (4) & w/o adapter pretraining
& 5.291 & 59.318 & 47.946 & 2.884 & 0.510 & 0.593 & 0.376
& 8.802 & 95.417 & 73.218 & 2.358 & 0.561 & 0.201 & 0.421 \\
& (5) & fixed adapter
& 5.452 & 59.984 & 48.713 & 2.971 & 0.526 & 0.609 & 0.389
& 9.087 & 96.936 & 75.021 & 2.451 & 0.583 & 0.212 & 0.438 \\
& (6) & joint optimization
& 5.408 & 59.762 & 48.531 & 2.943 & 0.521 & 0.604 & 0.384
& 9.014 & 96.573 & 74.612 & 2.427 & 0.578 & 0.209 & 0.432 \\
\midrule
\multicolumn{3}{c|}{Full Model (Transformer)}
& \best{5.861} & \best{61.069} & \best{50.581} & \best{3.179} & \best{0.564} & \best{0.643} & \best{0.425}
& \best{9.690} & \best{99.031} & \best{78.072} & \best{2.667} & \best{0.627} & \best{0.235} & \best{0.481} \\
\bottomrule
\end{tabular}%
}
\end{table*}

\begin{table}[t]
\centering
\caption{Comparison of runtime, computational cost (GFLOPs), and model size (Params) on IVF and MIF tasks. All values are averaged per image.}
\label{tab:complexity}
\resizebox{\linewidth}{!}{%
\begin{tabular}{lccc|lccc}
\toprule
\multicolumn{4}{c|}{\textbf{IVF}} & \multicolumn{4}{c}{\textbf{MIF}} \\ 
\cmidrule(lr){1-4} \cmidrule(lr){5-8}
\textbf{Method} & \textbf{Time (s)} & \textbf{GFLOPs (G)} & \textbf{Params (M)} &
\textbf{Method} & \textbf{Time (s)} & \textbf{GFLOPs (G)} & \textbf{Params (M)} \\
\midrule
EMMA         & 0.058  & 8.861    & 1.516   & EMMA          & 0.009  & 8.861    & 1.516 \\
Text-DiFuse  & 23.818 & 2742.5    & 119.460 & Text-DiFuse   & 24.424 & 2742.5   & 119.460 \\
Tc-MoA       & 0.543  & 61.000   & 340.580 & Tc-MoA        & 0.161  & 61.000   & 340.580 \\
ReFusion     & 0.026  & 98.344   & 1.486   & ReFusion      & 0.012  & 98.344   & 1.486 \\
C2RF         & 0.089  & 123.869  & 1.325   & C2RF          & 0.112  & 123.869  & 1.325 \\
GIF-Net      & 0.077  & 39.814   & 0.823   & GIF-Net       & 0.084  & 39.814   & 0.823 \\
SAGE         & 0.020  & 29.285   & 0.136   & CCF           & 31.876 & 1114.000 & 552.810 \\
Omni-Fuse    & 2.614  & 172.923  & 78.320  & Mask-Difuser  & 0.503  & 995.706  & 171.262 \\
CLDyN        & 0.272     & 174.061       & 0.46      & CLDyN         & 0.311     & 174.06       & 0.46  \\
ISFusion     & 0.382    & 273.472      & 1.372     & ISFusion      &  0.401      & 273.472      & 1.372  \\
\midrule
Ours(CNN)         & 0.013 & 3.284 & 0.067 & Ours(CNN)         & 0.010 & 3.284 & 0.067 \\
Ours(Mamba)       & 0.008 & 2.137 & 0.038 & Ours(Mamba)       & 0.006 & 2.137 & 0.038 \\
Ours(Transformer) & 0.028 & 6.912 & 0.142 & Ours(Transformer) & 0.023 & 6.912 & 0.142 \\
\bottomrule
\end{tabular}}
\end{table}

\textbf{(e) Effect of joint optimization.}
We also compare alternating optimization with joint optimization, where LFA and the fusion network are updated simultaneously. Joint optimization gives lower results than the full model. This suggests that changing the supervision space and the fused output at the same time can introduce unstable mutual feedback. Alternating optimization is more stable because one side is fixed while the other is updated.

\section{Model Complexity and Overhead Analysis}
Table~\ref{tab:complexity} compares the runtime, computational cost, and parameter size of representative IVF and MIF methods. Overall, our variants maintain a compact model scale and low computational overhead across both tasks. This indicates that the proposed framework does not rely on a heavy fusion backbone to achieve effective fusion performance.

Compared with many fusion methods, our models require fewer parameters and lower computation while keeping competitive inference speed. The three variants also provide different efficiency-performance trade-offs: the Mamba variant is the most lightweight, the CNN variant offers a balanced design, and the Transformer variant introduces a moderate increase in cost for stronger representation ability. These results show that the proposed relation-constrained supervision can work effectively with lightweight architectures and is suitable for practical fusion scenarios where efficiency matters.

\section{Implementation Details}
\label{app:Implementation Details}
\textbf{Feature extraction and adapter.}
We use frozen DINO and CLIP as feature providers, denoted by $\Psi$. For each input image $X$, we extract multi-level features from DINO layer 4, CLIP layer 6, and DINO layer 8, denoted by $f_D^4(X)$, $f_C^6(X)$, and $f_D^8(X)$, respectively. Each feature map is projected by a $1\times1$ convolution, followed by normalization and bilinear resizing. The corresponding projection operators are denoted by $P_D^4$, $P_C^6$, and $P_D^8$. The projected features are resized to a common token grid of $16\times16$ and projected to 256 channels. They are then flattened into token sequences and concatenated along the channel dimension. The feature aligner $G_{\mathrm{align}}$ is implemented as an MLP with hidden dimension 512 and output dimension 256, producing the aligned representation $y_X$.

In the relation reasoning module, the pair encoder $E_{\mathrm{pair}}$ takes $[y_A,y_B,|y_A-y_B|,y_A\odot y_B]$ as input and produces the pair feature $q_{A,B}$. The source-specific heads $H_A$ and $H_B$ produce the pair-conditioned source representations $z_A$ and $z_B$. The prediction heads $h_m$, $h_d$, and $h_r$ estimate sharedness $m_{A,B}$, dominance $d_{A,B}$, and coordination radius $r_{A,B}$, respectively. The pair encoder is implemented as an MLP with GELU activation and LayerNorm, and the prediction heads are lightweight linear layers. Sharedness and dominance are normalized by the sigmoid function, while the coordination radius is constrained by the softplus function.

\textbf{Hyperparameter settings.}
For the relation-aware source-pair score, we set the weighting coefficients to $\lambda_m=1.0$, $\lambda_d=0.5$, and $\lambda_r=0.5$. The numerical stability constant is set to $\epsilon=1\times10^{-6}$. The source-side ranking margin and fused-side ranking margin are set to $\delta_{\mathrm{src}}=0.2$ and $\delta_{\mathrm{fuse}}=0.2$, respectively. For the source-preservation update conflict analysis, the gradient weight is set to $\lambda_g=1.0$. We use patch size $16\times16$ and define the high-disagreement region $\Omega$ as the top 30\% patches ranked by source intensity and gradient discrepancy. $\alpha_{1}$ and $\alpha_{2}$ in $\mathcal{L}_{\mathrm{warm}}$ are set to 1 and 2.

\begin{figure}[t]
 \centering
  \includegraphics[width=1\textwidth]{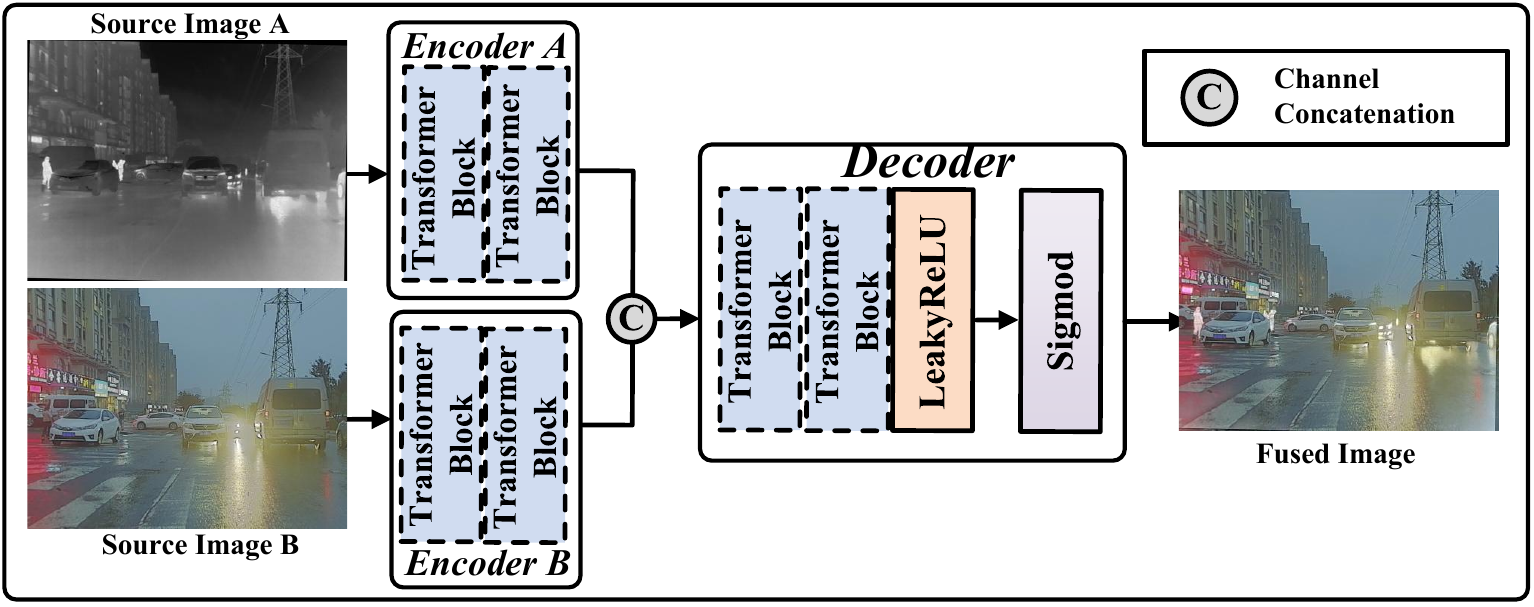}
  \caption{Schematic diagram of fusion network.}
  \label{fig:fusion-net}
\end{figure}

\textbf{Fusion network architecture.}
As shown in Fig.~\ref{fig:fusion-net}, our FusionNet adopts a two-branch encoder-decoder architecture. The two source images are first processed by two modality-specific encoders, denoted as Encoder A and Encoder B. These two encoders have the same architecture but do not share weights, allowing each branch to learn source-specific feature extraction. Taking the Transformer-based backbone as an example, each encoder is composed of stacked Transformer blocks and extracts a feature from its corresponding source image. The two latent features are then concatenated along the channel dimension and fed into the decoder. The decoder aggregates the concatenated feature through stacked Transformer blocks and a lightweight output head with LeakyReLU and Sigmoid, producing the final fused image.

Under this unified design, we instantiate Transformer-~\cite{zamir2022restormer}, CNN-~\cite{zhang2020ifcnn}, and Mamba-based~\cite{liu2024vmamba} fusion backbones to verify that the proposed supervision paradigm is not tied to a specific architecture. The Transformer variant is illustrated in Fig.~\ref{fig:fusion-net}. The CNN and Mamba variants follow the same two-branch encoder-decoder pipeline, and their only difference is that the Transformer blocks are replaced by CNN blocks~\cite{zhang2020ifcnn} and VMamba blocks~\cite{liu2024vmamba}, respectively.

\textbf{Architecture and Pretraining of the Shared Token Encoder.} 
$G_{\mathrm{shr}}$ is implemented as a lightweight token-wise MLP with shared weights across all modalities. Given an aligned representation $z\in\mathbb{R}^{N\times C}$, each token is processed independently as:
{\setlength{\abovedisplayskip}{3.5pt}%
\setlength{\belowdisplayskip}{3.5pt}%
\begin{equation}
G_{\mathrm{shr}}(z)
=
W_2\,\phi\!\left(W_1\,\mathrm{LN}(z)\right),
\end{equation}
}where $\mathrm{LN}(\cdot)$ denotes LayerNorm, $\phi(\cdot)$ denotes GELU, and $W_1$ and $W_2$ are linear projections. The output dimension is kept identical to that of $z$, allowing the shared representation to remain in the same supervision space. The same $G_{\mathrm{shr}}$ is applied to the two source representations and the fused representation. The corresponding residual representations $u_A$ and $u_B$ are defined by direct subtraction.

Following the correlation-driven feature decomposition strategy of CDDFuse~\cite{zhao2023cddfuse}, we pretrain $G_{\mathrm{shr}}$ using matched source pairs only. We encourage the shared representations $c_A$ and $c_B$ to be highly correlated, while reducing the correlation between the corresponding residual representations $u_A$ and $u_B$. We define the correlation between two token representations as:
{\setlength{\abovedisplayskip}{3.5pt}%
\setlength{\belowdisplayskip}{3.5pt}%
\begin{equation}
\mathrm{CC}(x,y)
=
\frac{
\left\langle x-\mu_x,\,y-\mu_y\right\rangle
}{
\|x-\mu_x\|_2\|y-\mu_y\|_2+\epsilon
},
\end{equation}
}where $\mu_x$ and $\mu_y$ denote the corresponding mean representations. Following this correlation-driven decomposition principle, the pretraining objective is defined as:
{\setlength{\abovedisplayskip}{3.5pt}%
\setlength{\belowdisplayskip}{3.5pt}%
\begin{equation}
\mathcal{L}_{G}
=
1-\mathrm{CC}(c_A,c_B)
+
\mathrm{CC}^{2}(u_A,u_B),
\end{equation}
}where the first term encourages $G_{\mathrm{shr}}$ to retain information consistently represented in both modalities, while the second term suppresses correlated information in the residual components. After pretraining, the parameters of $G_{\mathrm{shr}}$ are frozen. The same encoder is subsequently applied to $z_A$, $z_B$, and $z_F$ during relation-constrained supervision and source--fused ranking. 

\section{Limitations}
Although our method achieves strong fusion performance with a compact model design, it still has some limitations. First, the relation-constrained supervision relies on pretrained vision models to construct the supervision space. Its effectiveness may therefore be influenced by the representation quality and domain coverage of the selected pretrained models. Second, the current design mainly focuses on pairwise source relations. Extending it to more complex fusion settings with more than two modalities may require additional relation modeling. Future work will explore more robust supervision construction and broader adaptation to multi-modal and multi-source fusion scenarios.
\newpage
\section*{NeurIPS Paper Checklist}

\begin{enumerate}

\item {\bf Claims}
    \item[] Question: Do the main claims made in the abstract and introduction accurately reflect the paper's contributions and scope?
    \item[] Answer: \answerYes{} 
    \item[] Justification: See our abstract.

\item {\bf Limitations}
    \item[] Question: Does the paper discuss the limitations of the work performed by the authors?
    \item[] Answer: \answerYes{} 
    \item[] Justification: See Appendix.

\item {\bf Theory assumptions and proofs}
    \item[] Question: For each theoretical result, does the paper provide the full set of assumptions and a complete (and correct) proof?
    \item[] Answer: \answerNA{} 
    \item[] Justification: Our paper does not include theoretical results

    \item {\bf Experimental result reproducibility}
    \item[] Question: Does the paper fully disclose all the information needed to reproduce the main experimental results of the paper to the extent that it affects the main claims and/or conclusions of the paper (regardless of whether the code and data are provided or not)?
    \item[] Answer: \answerYes{} 
    \item[] Justification: We provide full details for reproducing the proposed method.

\item {\bf Open access to data and code}
    \item[] Question: Does the paper provide open access to the data and code, with sufficient instructions to faithfully reproduce the main experimental results, as described in supplemental material?
    \item[] Answer: \answerYes{} 
    \item[] Justification: Our code and data will be publicly available.

\item {\bf Experimental setting/details}
    \item[] Question: Does the paper specify all the training and test details (e.g., data splits, hyperparameters, how they were chosen, type of optimizer) necessary to understand the results?
    \item[] Answer: \answerYes{} 
    \item[] Justification: We provide complete training and testing settings in Section 4 and Appendix.

\item {\bf Experiment statistical significance}
    \item[] Question: Does the paper report error bars suitably and correctly defined or other appropriate information about the statistical significance of the experiments?
    \item[] Answer: \answerNA{} 
    \item[] Justification: See it in Appendix.

\item {\bf Experiments compute resources}
    \item[] Question: For each experiment, does the paper provide sufficient information on the computer resources (type of compute workers, memory, time of execution) needed to reproduce the experiments?
    \item[] Answer: \answerYes{} 
    \item[] Justification: See Appendix.
    
\item {\bf Code of ethics}
    \item[] Question: Does the research conducted in the paper conform, in every respect, with the NeurIPS Code of Ethics \url{https://neurips.cc/public/EthicsGuidelines}?
    \item[] Answer: \answerYes{} 
    \item[] Justification: The paper conducted in this paper complies with the NeurIPS Code of Ethics.

\item {\bf Broader impacts}
    \item[] Question: Does the paper discuss both potential positive societal impacts and negative societal impacts of the work performed?
    \item[] Answer: \answerYes{} 
    \item[] Justification: See it in appendix.
    
\item {\bf Safeguards}
    \item[] Question: Does the paper describe safeguards that have been put in place for responsible release of data or models that have a high risk for misuse (e.g., pre-trained language models, image generators, or scraped datasets)?
    \item[] Answer: \answerNA{} 
    \item[] Justification: There is no such risks in our paper.

\item {\bf Licenses for existing assets}
    \item[] Question: Are the creators or original owners of assets (e.g., code, data, models), used in the paper, properly credited and are the license and terms of use explicitly mentioned and properly respected?
    \item[] Answer: \answerYes{} 
    \item[] Justification: In Section 4, we introduce all datasets and metrics we used.

\item {\bf New assets}
    \item[] Question: Are new assets introduced in the paper well documented and is the documentation provided alongside the assets?
    \item[] Answer: \answerNA{} 
    \item[] Justification: Our paper does not release new assets.

\item {\bf Crowdsourcing and research with human subjects}
    \item[] Question: For crowdsourcing experiments and research with human subjects, does the paper include the full text of instructions given to participants and screenshots, if applicable, as well as details about compensation (if any)? 
    \item[] Answer: \answerNA{} 
    \item[] Justification: Our paper does not involve crowdsourcing nor research with human subjects.

\item {\bf Institutional review board (IRB) approvals or equivalent for research with human subjects}
    \item[] Question: Does the paper describe potential risks incurred by study participants, whether such risks were disclosed to the subjects, and whether Institutional Review Board (IRB) approvals (or an equivalent approval/review based on the requirements of your country or institution) were obtained?
    \item[] Answer: \answerNA{} 
    \item[] Justification: Our paper does not involve crowdsourcing nor research with human subjects.

\item {\bf Declaration of LLM usage}
    \item[] Question: Does the paper describe the usage of LLMs if it is an important, original, or non-standard component of the core methods in this research? Note that if the LLM is used only for writing, editing, or formatting purposes and does \emph{not} impact the core methodology, scientific rigor, or originality of the research, declaration is not required.
    \item[] Answer: \answerNA{} 
    \item[] Justification: The core method development in this research does not involve LLMs as any important, original, or non-standard components.

\end{enumerate}
\end{document}